\documentclass[final,5p,times,twocolumn]{elsarticle}

\usepackage{amssymb}
\usepackage{amsmath}
\usepackage{hyperref}
\usepackage{float}
\usepackage{graphicx}
\usepackage{algorithm}
\usepackage{algpseudocode} 
\usepackage{amsmath,amssymb}
\usepackage{booktabs}
\usepackage{multirow}
\usepackage{colortbl}
\usepackage{pifont}  
\usepackage{caption}
\usepackage{subcaption}  
\usepackage{float}
\usepackage{adjustbox}  
\usepackage[table,xcdraw]{xcolor}
\usepackage{ulem}
\usepackage{cleveref}
\usepackage{ragged2e}
\usepackage{afterpage} 
\usepackage{placeins}
\definecolor{yellow}{RGB}{255,192,0}  
\definecolor{red}{RGB}{180,0,0}   
\definecolor{blue_mm}{RGB}{0,112,192} 
\definecolor{green}{RGB}{0,150,0} 
\definecolor{lightred}{RGB}{255,0,0} 
\definecolor{deepgreen}{RGB}{50,118,64} 

\Crefname{figure}{Fig.}{Figs.}

\usepackage{soul, color, xcolor}
\definecolor{myyellow}{RGB}{255,255,0}
\sethlcolor{myyellow}  
\newif\ifhighlight
\highlightfalse     
\newcommand{\hla}[1]{\ifhighlight\hl{#1}\else#1\fi}

\soulregister{\textbf}7
\soulregister{\textit}7
\soulregister{\cite}7 
\soulregister{\cref}7 
\soulregister{\Cref}7  
\soulregister{\underline}7 %
\soulregister{\url}7 
\soulregister{\textit}{1}

\journal{Knowledge-Based Systems}

\begin{document}

\begin{frontmatter}



\title{Towards Robust Infrared Small Target Detection: A Feature-Enhanced and Sensitivity-Tunable Framework}


\author{Jinmiao Zhao$^a$$^{,b}$$^{,c}$, Zelin Shi$^{b*}$, Chuang Yu$^a$$^{,b}$$^{,c}$, Yunpeng Liu$^b$, Yimain Dai$^{d}$} 

\affiliation[a]{organization={Key Laboratory of Opto-Electronic Information Processing, Chinese Academy of Sciences}, city={Shenyang},country={China}}
\affiliation[b]{organization={Shenyang Institute of Automation, Chinese Academy of Sciences}, city={Shenyang},country={China}}
\affiliation[c]{organization={University of Chinese Academy of Sciences}, city={Beijing},country={China}}
\affiliation[d]{organization={Nankai University}, city={Tianjing},country={China}}

\begin{abstract}
Recently, single-frame infrared small target (SIRST) detection technology has attracted widespread attention. Different from most existing deep learning-based methods that focus on improving network architectures, we propose a feature-enhanced and sensitivity-tunable (FEST) framework, which is compatible with existing SIRST detection networks and further enhances their detection performance. The FEST framework improves the model's robustness from two aspects: feature enhancement and target confidence regulation. \hla{For feature enhancement, we employ a multi-scale fusion strategy to improve the model’s perception to multi-scale features of multi-size targets, and design an edge enhancement difficulty mining (EEDM) loss to guide the network to continuously focus on challenging target regions and edge features during training. For target confidence regulation, an adjustable sensitivity (AS) strategy is proposed for network post-processing. This strategy enhances the model’s adaptability in complex scenarios and significantly improves the detection rate of infrared small targets while maintaining segmentation accuracy. Extensive experimental results show that our FEST framework can effectively enhance the performance of existing SIRST detection networks. The code is available at \url{https://github.com/YuChuang1205/FEST-Framework}}.
\end{abstract}




\end{frontmatter}



\section{Introduction}
\label{sec: introduction}
Infrared imaging technology, with its excellent low-light imaging capabilities, demonstrates unique advantages in \hla{terms of} complex environment perception, driving the widespread application of infrared detection technologies~\cite{han2022kcpnet,gao2024defense,terren2023detection,yu2024easy}. Among them, SIRST detection, as a fundamental technology, plays an important role in many fields, including target recognition and tracking~\cite{huang2023anti,ying2025infrared}, early warning systems~\cite{xu2023multiscale,zhao2022single}, traffic surveillance~\cite{ying2025visible,odat2017vehicle}, and maritime search and rescue~\cite{zhao2024infrared,zhang2022rkformer}.

The research on SIRST detection can be divided into non-deep learning-based methods and deep learning-based methods. Non-deep learning-based SIRST detection methods mainly use the characteristics of infrared small target images for detection~\cite{arce1987theoretical,tomasi1998bilateral,li2010robust,chen2014novel,chen2013local,wei2016multiscale,gao2013infrared,dai2017non}. These methods often employ traditional image processing techniques, including filtering, sparse representation, low-rank decomposition, and background modeling, to enhance target features while reducing background interference. They are usually optimized for specific scenarios and can achieve good detection results under specific imaging conditions. However, owing to the complexity and variability of practical scenarios, these traditional methods often suffer from many false detections or even failures. With the rapid development of deep learning, many researchers have focused on leveraging the powerful representation capabilities of deep neural networks to extract richer, more efficient, and more robust features, \hla{which have been widely applied across various research fields~\cite{yu2023feature,tian2018detection,yu2022precise,zhang2023motor,raja2025bayesian}}. In light of current technological trends and growing application demands, deep learning-based SIRST methods~\cite{dai2021asymmetric,dai2021attentional,yu2022infrared,yu2022pay,li2022dense,wu2022uiu,zhao2023gradient,zhao2024multi,zhang2022isnet} have achieved remarkable progress. This type of method usually uses deep learning models \hla{such as convolutional neural networks (CNNs)~\cite{lecun1998gradient} and Vision Transformers (ViT)~\cite{dosovitskiy2020image} to automatically learn the spatial and textural features of the target from the data, thereby improving detection performance.} However, most current methods mainly focus on optimizing the network structure while ignoring other aspects that may improve the detection performance. Therefore, it is of great significance to design an efficient and robust SIRST detection framework to enhance the overall performance of networks in complex environments.

Different from most existing deep learning-based methods that focus on improving network architectures, we propose a feature-enhanced and sensitivity-tunable (FEST) framework. The framework can be flexibly compatible with existing SIRST networks, thereby improving its robustness in complex scenarios. Specifically, in the aspect of the feature enhancement, a multi-scale fusion strategy is employed to improve the model’s perception to multi-scale features of multi-size targets. \hla{By integrating information across multiple spatial scales, the framework enables comprehensive multi-scale feature representation of a single target and enhances adaptability to scale variations among multiple targets.} In addition, to guide the network to pay more attention to challenging target regions and edge features during training, we propose an edge enhancement difficulty mining (EEDM) loss. This loss helps the model learn valuable features more effectively and greatly avoids the risk of small targets being submerged by the background. In the aspect of the target confidence regulation, to further enhance the adaptability of our framework to different scenarios, we propose an adjustable sensitivity (AS) strategy for post-processing. This strategy introduces the concepts of strong targets and weak targets. It can significantly improve the detection rate of infrared small targets while ensuring a small change in the false alarm rate and segmentation accuracy, thus having important practical application value. Extensive experimental results show that the proposed FEST framework can significantly enhance the performance of existing SIRST detection networks. \hla{Notably, the multi-scale direction-aware network (MSDA-Net) equipped with the FEST framework won \textit{the first prize in the “PRCV 2024 wide-area infrared small target detection competition”}.} The contributions of this manuscript can be summarized as follows:
\begin{itemize}
    \item A feature-enhanced and sensitivity-tunable (FEST) framework is constructed, which can be flexibly compatible with the existing SIRST networks and effectively enhances their detection performance.
    \item For feature enhancement, a multi-scale fusion strategy is adopted to improve the model's perception to multi-scale features of multi-size targets. In addition, an edge enhancement difficulty mining loss is constructed, which can guide the network to pay more attention to edge features and challenging target regions.
    \item For target confidence regulation, an adjustable sensitivity strategy is proposed, which deeply mines the generated probability mask information and innovatively defines the concepts of strong targets and weak targets. This strategy is conducive to significantly improving the detection rate while ensuring the false alarm rate and segmentation accuracy.
\end{itemize}

\section{Related Work}
\label{sec:related-work}
\subsection{Non-deep learning-based SIRST detection methods}
Non-deep learning infrared small target detection methods can be roughly divided into three categories: background suppression-based methods, human visual system-based methods and image data structure-based methods. Among them, background suppression-based methods include spatial domain filtering (e.g., \hla{the} Max-Median filter~\cite{arce1987theoretical} and two-dimensional minimum mean square error filter~\cite{tomasi1998bilateral}) and transform domain filtering (e.g., \hla{the} dual-tree complex wavelet~\cite{li2010robust} and bidimensional empirical mode decomposition (BEMD)~\cite{chen2014novel}). Spatial domain filtering methods are computationally efficient but suffer performance degradation in complex scenes due to the difficulty in predefining effective templates. In contrast, transform domain methods offer stronger background modeling capabilities but incur high computational costs, limiting their practical applicability. \hla{Human visual system-based methods} simulate the human eye's perception of local contrast to detect salient objects in images. \hla{Representative methods} include the local contrast measure (LCM)~\cite{chen2013local} and multi-scale patch-based contrast measure (MPCM)~\cite{wei2016multiscale}. These methods can effectively suppress large-area bright background interference and improve target saliency. However, they still tend to produce many false detections when strong noise or interference is present in the image. Image data structure-based methods formulate the detection task as a low-rank and sparse matrix separation problem, leveraging the non-local self-similarity of the background and the sparsity of the target to achieve foreground-background separation. \hla{Representative methods} include the infrared patch-image model (IPI)~\cite{gao2013infrared} and the non-negative infrared patch-image model based on partial sum minimization of singular values (NIPPS)~\cite{dai2017non}. These methods generally offer effective separation performance but remain sensitive to strong edges in the background and are computationally expensive.

\begin{figure*}[t]
        \vspace{0pt}
	\centering
	\includegraphics[width=1\textwidth]{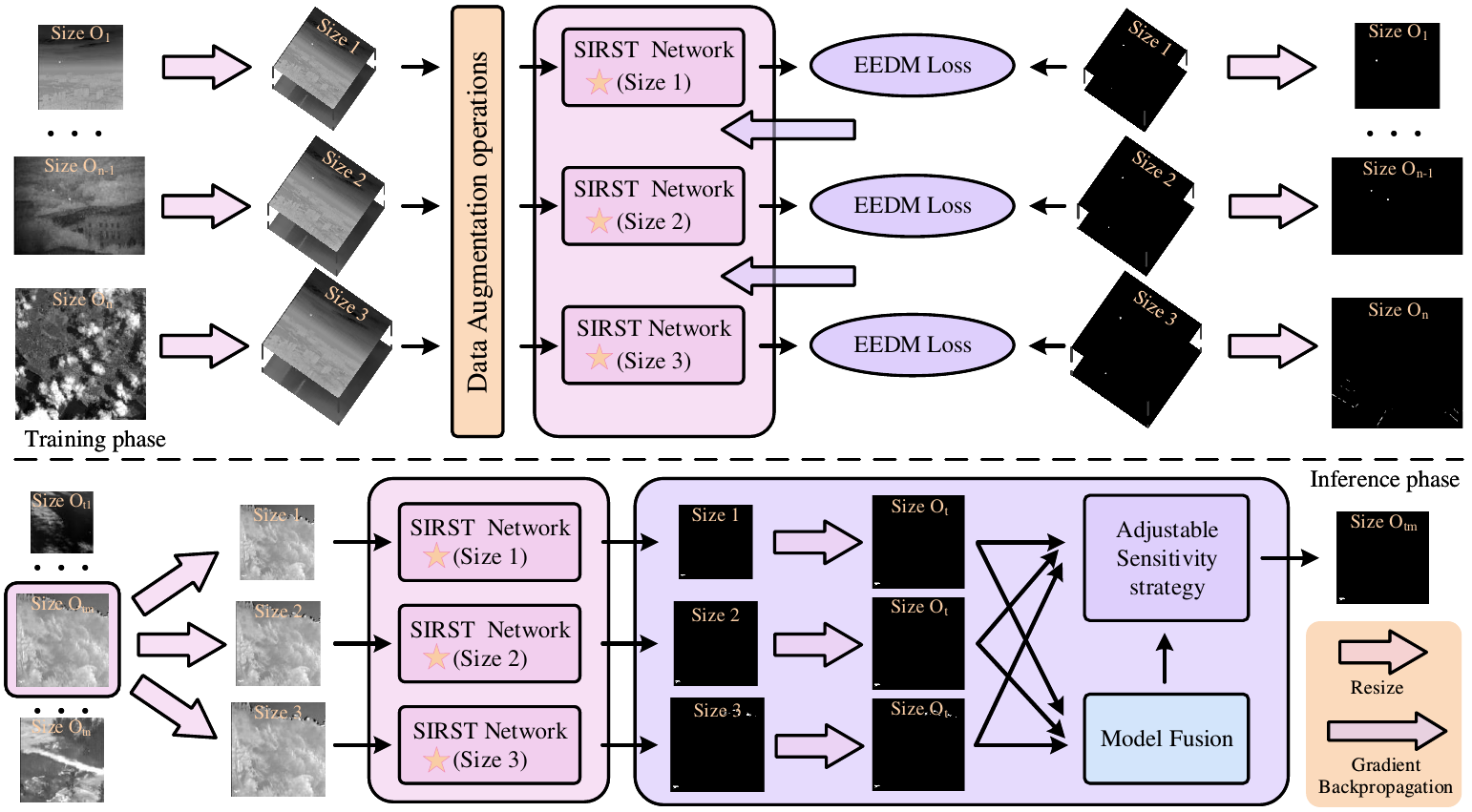}
	\vspace{-15pt}
	\caption{\hla{Overall structure of the FEST framework. ``SIRST Network'' denotes the selected existing SIRST networks. Size $O_1$, ..., Size $O_{n-1}$, and Size $O_n$ denote the actual sizes of the original training images. Size 1, Size 2, and Size 3 denote the input resolutions corresponding to each model. Size $O_{t1}$, ..., Size $O_{t(m-1)}$, and Size $O_{tm}$ denote the actual sizes of the original test images.}}
	\label{fig:fig_method_fest}
	\vspace{-3pt}
\end{figure*}

\subsection{Deep learning-based SIRST detection methods}
Different from non-deep learning-based methods, deep learning-based methods learn the features of infrared small targets in a data-driven manner~\cite{zhao2024refined,yu2024lr}. Specifically, Dai et al. propose the asymmetric context module (ACM)~\cite{dai2021asymmetric} and the attention local contrast network (ALCNet)~\cite{dai2021attentional}, which for the first time introduced the local contrast idea in traditional methods into the deep learning-based method. To further explore the relationship between local areas, Yu et al. successively proposed the multi-scale local contrastive learning network (MLCL-Net)~\cite{yu2022infrared} and the improved local contrastive learning network (ALCL-Net)~\cite{yu2022pay}, which enhanced the network's perception ability of infrared small targets. Subsequently, Li et al. proposed a densely nested attention network (DNANet)~\cite{li2022dense} based on the UNet++ architecture , aiming to alleviate the problem of target information loss caused by downsampling operations in infrared small target detection. At the same time, Wu et al. proposed a U-Net nested network (UIU-Net)~\cite{wu2022uiu}, which achieves richer multi-scale feature expression by nesting a lightweight U-Net network in the backbone U-Net. In recent years, Zhao et al. further emphasize the gradient information and directional characteristics of infrared small target images and propose GGL-Net~\cite{zhao2023gradient} and MSDA-Net~\cite{zhao2024multi} respectively. GGL-Net introduces amplitude images at multiple stages of network training, prompting the network to pay more attention to discriminative features. MSDA-Net further promotes the refinement of infrared small target segmentation by emphasizing the attention to multi-directional high-frequency features. In addition, Zhang et al. proposed an infrared shape network (ISNet)~\cite{zhang2022isnet} to improve the accuracy of target recognition by emphasizing the extraction of shape features in infrared images. However, most current methods focus on optimizing the network structure while ignoring other aspects that may improve the detection performance. Therefore, we aim to build an efficient and robust SIRST detection framework to further improve the performance of existing SIRST detection networks.

\section{Method}
\label{sec:method}
\subsection{Overall Framework}
From \Cref{fig:fig_method_fest}, the proposed scheme consists of two parts: the model training phase and the model inference phase. In the model training phase, first, we perform multi-scale transformations on the original training images, adjusting them to multiple resolutions through scaling operations. Then, the images at different scales are individually fed into the corresponding basic models for training. The basic models can be any existing SIRST \hla{networks}. Finally, to further improve the detection accuracy of the network, we use the proposed EEDM \hla{loss} as the loss function to optimize the network. In the model inference phase, first, we perform multi-scale transformations on the original test images and input the transformed images into the corresponding multi-scale models for inference. Next, the feature maps generated by multiple scale models are uniformly resampled to the original test image size to ensure feature alignment. Then, the multi-scale fusion strategy is used to integrate the prediction results of multiple scale models. This strategy can improve the model's perception of different target scales and enhance its robustness in complex environments. Finally, the proposed AS strategy is used to post-process the fused results. This strategy can significantly improve the detection rate of infrared small targets while ensuring the false alarm rate and segmentation accuracy. \hla{To further illustrate the implementation of the FEST framework, the training and inference phases are presented in Algorithm 1 and Algorithm 2.}

\renewcommand{\algorithmicrequire}{\textbf{Input:}}
\renewcommand{\algorithmicensure}{\textbf{Output:}}
\renewcommand{\thealgorithm}{1}
\begin{algorithm}[!t]
\caption{\hla{FEST Framework (Training phase)}}
\label{alg:training_phase}
\begin{algorithmic}[1] 
\Require Resize set $\{r_1, r_2, r_3\}$; Batch infrared images $I_{i = 1}^{{N_b}}$; Batch ground-truth masks $M_{i = 1}^{{N_b}}$; Network $\varphi$ (including ${\varphi_{r1}}$, ${\varphi_{r2}}$, ${\varphi_{r3}}$ for different input image sizes); Learning rate $l_r$.
\Ensure Updated parameters $\{ {\Theta _{{\rm{r1}}}},{\Theta _{r2}},{\Theta _{r3}}\}$

\State \textbf{Initialize:} Give the initial value to ${\varphi_{r1}}$, ${\varphi_{r2}}$, ${\varphi_{r3}}$, and $l_r$.
\State \textbf{Step 1}: For each batch of data $\{(I, M)\} _{i = 1}^{{N_b}}$, resize it to $\{r_1, r_2, r_3\}$:
\[\{({I_{r1}},{M_{r1}})\} _{i = 1}^{{N_b}}\; \leftarrow \;{Re_{r1}}(\{(I,M)\}_{i = 1}^{{N_b}})\]
\[\{ ({I_{r2}},{M_{r2}})\} _{i = 1}^{{N_b}}\; \leftarrow \;{Re_{r2}}(\{(I,M)\}_{i = 1}^{{N_b}})\]
\[\{ ({I_{r3}},{M_{r3}})\} _{i = 1}^{{N_b}}\; \leftarrow \;{Re_{r3}}(\{(I,M)\}_{i = 1}^{{N_b}})\]
\State \textbf{Step 2}: For each batch of data with different input sizes $\{ ({I_{r}},{M_{r}})\} _{i = 1}^{{N_b}}$, data augmentation is performed to obtain $\{({I'_r},{M'_r})\} _{i = 1}^{{N_b}}$.

\State \textbf{Step 3}: Feed each batch of the infrared images $\{ \{ {I'_{r1}}\}_{i = 1}^{{N_b}},\{ {I'_{r2}}\} _{i = 1}^{{N_b}},\{ {I'_{r3}}\}_{i = 1}^{{N_b}}\}$, into the corresponding network to obtain prediction maps $\{\{ {P_{r1}}\} _{i = 1}^{{N_b}},\{ {P_{r2}}\} _{i = 1}^{{N_b}},\{ {P_{r3}}\} _{i = 1}^{{N_b}}\}$:
\[\{ {P_{r1}}\} _{i = 1}^{{N_b}}\; \leftarrow \;{\varphi _{r1}}(\{ {I_{r1}}\} _{i = 1}^{{N_b}})\]
\[\{ {P_{r2}}\} _{i = 1}^{{N_b}}\; \leftarrow \;{\varphi _{r2}}(\{ {I_{r2}}\} _{i = 1}^{{N_b}})\]
\[\{ {P_{r3}}\} _{i = 1}^{{N_b}}\; \leftarrow \;{\varphi _{r3}}(\{ {I_{r3}}\} _{i = 1}^{{N_b}})\]

\State \textbf{Step 4}: Based on formulas (1)–(6), compute the EEDM loss using the prediction maps $\{\{ {P_{r1}}\} _{i = 1}^{{N_b}},\{ {P_{r2}}\} _{i = 1}^{{N_b}},\{ {P_{r3}}\} _{i = 1}^{{N_b}}\}$ and the ground-truth masks $\{ \{ {M'_{r1}}\} _{i = 1}^{{N_b}},\{ {M'_{r2}}\} _{i = 1}^{{N_b}},\{ {M'_{r3}}\} _{i = 1}^{{N_b}}\}$ to obtain $L_{r1}$, $L_{r2}$, $L_{r3}$. 

\State \textbf{Step 5}: Based on the optimizer, update the network parameters $\{ {\Theta _{{\rm{r1}}}},{\Theta _{r2}},{\Theta _{r3}}\}$:
\[{\Theta _{r1}}\; \leftarrow \;{\Theta _{r1}}\; - \;l_r{\nabla _{{\Theta _{r1}}}}{L_{r1}}\]
\[{\Theta _{r2}}\; \leftarrow \;{\Theta _{r2}}\; - \;l_r{\nabla _{{\Theta _{r2}}}}{L_{r2}}\]
\[{\Theta _{r3}}\; \leftarrow \;{\Theta _{r3}}\; - \;l_r{\nabla _{{\Theta _{r3}}}}{L_{r3}}\]

\State \textbf{Step 6}: Repeat Step 1-5 until the training is completed.

\State \textbf{Return:} ${\Theta _{{\rm{r1}}}}$, ${\Theta _{r2}}$, ${\Theta _{r3}}$.

\end{algorithmic} 
\end{algorithm}

\subsection{Multi-scale fusion strategy}
In the SIRST detection task, the target scale is affected by factors such as \hla{the} imaging scene, shooting distance and sensor equipment, and can usually be divided into point targets, spot targets and extended targets~\cite{yu2024easy,zhao2024multi}. In complex scenes, a single-scale network model struggles to handle small targets of various sizes simultaneously, often resulting in missed or false detections. Additionally, considering that this study aims to build a detection framework compatible with the existing SIRST \hla{networks}, we seek to enhance \hla{the} detection performance without altering the network structure. Therefore, we adopt a multi-scale fusion strategy to enhance the network's adaptability to different scales targets by integrating feature information from multiple scale models, effectively alleviating the shortcomings of a single-scale model.

From \Cref{fig:fig_method_fest}, the multi-scale fusion strategy is reflected in three parts. First, in the model training phase, without modifying the structure of the SIRST network, images of different sizes are used for training to obtain models suitable for different resolutions. Secondly, in the inference phase, the original input image is transformed into multiple sizes and fed into the corresponding models for inference. Finally, to fully leverage the detection advantages of models at different sizes, the prediction results are resampled back to the original image size and integrated to produce the detection result.

The multi-scale model fusion strategy is conducive to fully mining the semantic features and detail features of multi-size targets at different scales, making the model more accurate and reliable in identifying and locating targets. In addition, practical detection scenarios are often more complex and changeable, so fusing models of different scales can help improve the \hla{model’s} robustness in complex backgrounds. In practical applications, the fusion scales and the number of models can be reasonably chosen according to specific task requirements to achieve an optimal balance between detection performance and inference efficiency.

\subsection{Edge enhancement difficulty mining loss}
\begin{figure*}[t]
        \vspace{0pt}
	\centering
	\includegraphics[width=0.95\linewidth]{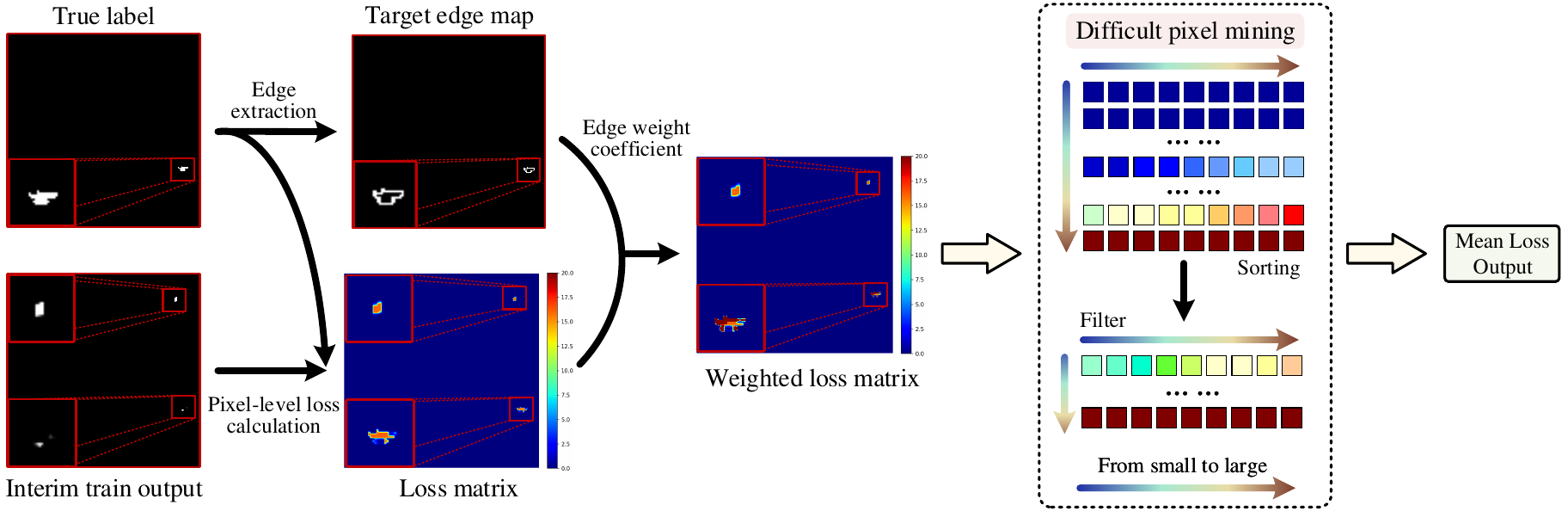}
	\vspace{-2pt}
	\caption{\hla{Illustration of Edge enhancement difficulty mining (EEDM) loss.}}
	\label{fig:fig_method_eedm}
	\vspace{-2pt}
\end{figure*}

\hla{In SIRST detection}, the small target area and lack of intrinsic features make it difficult to accurately locate in the image. Existing studies have \hla{demonstrated} that paying attention to edge information in this task can bring significant performance improvements~\cite{zhao2023gradient,zhang2022isnet}. Therefore, on the basis of the characteristics of infrared small targets, we design an EEDM loss. This loss can effectively guide the network to pay more attention to edge features and challenging target regions during training, thereby enhancing the model’s boundary awareness and improving its ability to learn from challenging samples. \hla{The workflow of the EEDM loss is shown in \Cref{fig:fig_method_eedm}.}

The EEDM loss consists of two components: edge pixel enhancement and difficult pixel mining. Edge pixel enhancement is performed preferentially. \hla{Specifically, the pixel-wise loss values are first computed using the binary cross-entropy loss~\cite{li2024rediscovering}, yielding the loss matrix ${L_{BCE}}$. At the same time, we perform edge extraction on the true label to obtain the target edge map. Based on this, an edge weighting coefficient $w$ is introduced to weight the target edge map and the loss matrix ${L_{BCE}}$, resulting in the weighted loss matrix ${L^*}$.} The formulas are as follows:
\begin{gather}
{L_{BCE}} = -[{y_i}\log ({\hat y_i}) + (1 - {y_i})\log (1 - {\hat y_i})]  \\[3pt]
e' = \left\{ {\begin{array}{*{20}{l}}
1 & \text{if}\; e = 0 \\
w & \text{if}\; e > 0
\end{array}} \right.  \\[3pt]
{L^*} = e' \times {L_{BCE}} 
\end{gather}
\hla{where ${y_i}$ denotes the true label of the $i_{th}$ pixel, ${\hat y_i}$ denotes the probability of the $i_{th}$ pixel being predicted as the target, $e$ denotes the target edge map, $w$ denotes the edge weighting coefficient, and $e'$ denotes the edge weighting map}.

The difficult pixel mining is performed subsequently. Specifically, first, the loss values corresponding to each pixel in ${L^*}$ are sorted. Secondly, pixels with loss values greater than or equal to the difficult pixel mining ratio are selected as the set of difficult \hla{loss values} ${L_{{\rm{hard}}}}$. Finally, the average of the difficult \hla{loss values} is computed to obtain the final loss value. The formulas are as follows:
\begin{gather}
{L_{sorted}} = sort({L^*})  \\[3pt]
{L_{{\rm{hard}}}} = \left\{ {{L_i}\mid {L_i} \ge {L_{{\rm{sorted}}}}\left[ {\left\lfloor {p \times N} \right\rfloor } \right]} \right\}  \\[3pt]
{L_{{\rm{EEDM}}}} = \frac{1}{{|{L_{{\rm{hard}}}}|}}\sum\limits_{i \in {L_{{\rm{hard}}}}} {{L_{hard}}} [i]
\end{gather}
\hla{where $\text{sort}(\cdot)$ denotes the sort operation, ${L_{sorted}}$ denotes the sorted sequence of all pixel-wise loss values, $N$ denotes the total number of pixel-wise loss values, $p$ denotes the difficult pixel mining ratio and ${L_{EEDM}}$ denotes the final output.}

On the one hand, EEDM loss uses edge information as an additional constraint by assigning higher weights to the target boundary regions, thereby enhancing the model's sensitivity to target boundaries. On the other hand, it dynamically discards a certain proportion of simple samples by difficult pixel mining, which enables the network to focus more on challenging regions and promotes the network to learn more discriminative features.

\renewcommand{\algorithmicrequire}{\textbf{Input:}}
\renewcommand{\algorithmicensure}{\textbf{Output:}}
\renewcommand{\thealgorithm}{2}
\begin{algorithm}[!t]
\caption{\hla{FEST Framework (Inference phase)}}
\label{alg:inference_phase}
\begin{algorithmic}[1] 
\Require Resize set $\{r_1, r_2, r_3\}$; Batch infrared images $I_{i = 1}^{{N_b}}$; Networks $\varphi  = \{ {\varphi _{r1}},{\varphi _{r2}},{\varphi _{r3}}\} $
\Ensure Predicted mask $\{ M\} _{i = 1}^{{N_b}}$

\State \textbf{Initialize:} Give the trained weights to $\varphi _{r1}$, $\varphi _{r2}$, $\varphi _{r3}$.

\State \textbf{Step 1}: For each batch of input images $\{ I\} _{i = 1}^{{N_b}}$, resize inputs into $\{{I_{r1}},{I_{r2}},{I_{r3}}\} _{i = 1}^{{N_b}}$

\State \textbf{Step 2}: Feed each resized infrared image batch $\{ \{ {I_{r1}}\} _{i = 1}^{{N_b}},\{ {I_{r2}}\} _{i = 1}^{{N_b}},\{ {I_{r3}}\} _{i = 1}^{{N_b}}\}$ into the corresponding networks to obtain prediction maps $\{ \{ {P_{r1}}\} _{i = 1}^{{N_b}},\{ {P_{r2}}\} _{i = 1}^{{N_b}},\{ {P_{r3}}\} _{i = 1}^{{N_b}}\}$:
\[\{ {P_{r1}}\} _{i = 1}^{{N_b}}\; \leftarrow \;{\varphi _{r1}}(\{ {I_{r1}}\} _{i = 1}^{{N_b}})\]
\[\{ {P_{r2}}\} _{i = 1}^{{N_b}}\; \leftarrow \;{\varphi _{r2}}(\{ {I_{r2}}\} _{i = 1}^{{N_b}})\]
\[\{ {P_{r3}}\} _{i = 1}^{{N_b}}\; \leftarrow \;{\varphi _{r3}}(\{ {I_{r3}}\} _{i = 1}^{{N_b}})\]

\State \textbf{Step 3}: Apply the multi-model fusion strategy on the prediction maps $\{ \{ {P_{r1}}\} _{i = 1}^{{N_b}},\{ {P_{r2}}\} _{i = 1}^{{N_b}},\{ {P_{r3}}\} _{i = 1}^{{N_b}}\}$ to obtain the fused map ${P_{fuse}}$:
\[{P_{fuse}}\; \leftarrow \;Mean(Re(\{{P_{r1}}\} _{i = 1}^{{N_b}}),\;Re(\{ {P_{r2}}\} _{i = 1}^{{N_b}}),\;Re(\{ {P_{r3}}\} _{i = 1}^{{N_b}}))\]

\State \textbf{Step 4}: Based on the prediction maps $\{ {P_{r1}}\} _{i = 1}^{{N_b}}$, $\{ {P_{r2}}\} _{i = 1}^{{N_b}}$, $\{ {P_{r3}}\} _{i = 1}^{{N_b}}$ and the fused map ${P_{fuse}}$, apply the proposed AS strategy to generate the final detection mask $\{ M\} _{i = 1}^{{N_b}}$

\State \textbf{Return:} $\{ M\} _{i = 1}^{{N_b}}$

\end{algorithmic} 
\end{algorithm}

\subsection{Adjustable sensitivity strategy}
\hla{In the small target segmentation task}, the network outputs a probability mask, \hla{where each pixel value} denotes its confidence of belonging to the target. Existing studies typically use a single threshold (0.5) to separate targets from the background~\cite{yu2024easy,zhao2024multi}. However, we believe that the inherent idea of directly using a single threshold cannot fully mine and utilize the information in the probability mask. Therefore, we propose an AS strategy aimed at deeply mining the information in the probability mask. Based on this, we innovatively define the concepts of strong targets and weak targets. Strong targets refer to those that are easy for the network to detect and have high confidence. These targets typically display more distinct features in the image, enabling the network to quickly identify them through learning. For strong targets, our goal is to achieve complete and precise segmentation. In contrast, weak targets refer to those that are difficult for the network to detect and have lower confidence. These targets may be challenging for the network to recognize \hla{because of} their shape or the complexity of the background. For weak targets, our goal is to identify them to meet the needs of practical scenarios.

Through careful observation of the probability mask generated by deep learning methods in this task, we interestingly \hla{find} that, influenced by data ambiguity and the network's feature extraction mechanism, the network's predicted probability mask for the target region exhibits a Gaussian-like distribution. Specifically, the probability is higher in the center region and gradually decreases \hla{toward} the edge, and finally approaches zero. Based on this observation and to better detect strong and weak targets, we design two thresholds $th1$ and $th2$ ($th1 > th2$) to adjust the model’s responsiveness to regions with different confidence levels. Specifically, when the network's confidence in a region is higher than $th1$, the region is defined as a strong target region. In this case, the network demonstrates high detection capability and can provide relatively accurate segmentation results. Additionally, when the confidence is between $th1$ and $th2$, the region is defined as a weak target region. For these weak target regions, although the network’s detection accuracy is low, reasonable post-processing can still help identify partial features of these targets to meet the detection needs of weak target detection in practical scenarios.

\begin{figure}[t]
        \vspace{0pt}
	\centering
	\includegraphics[width=\columnwidth]{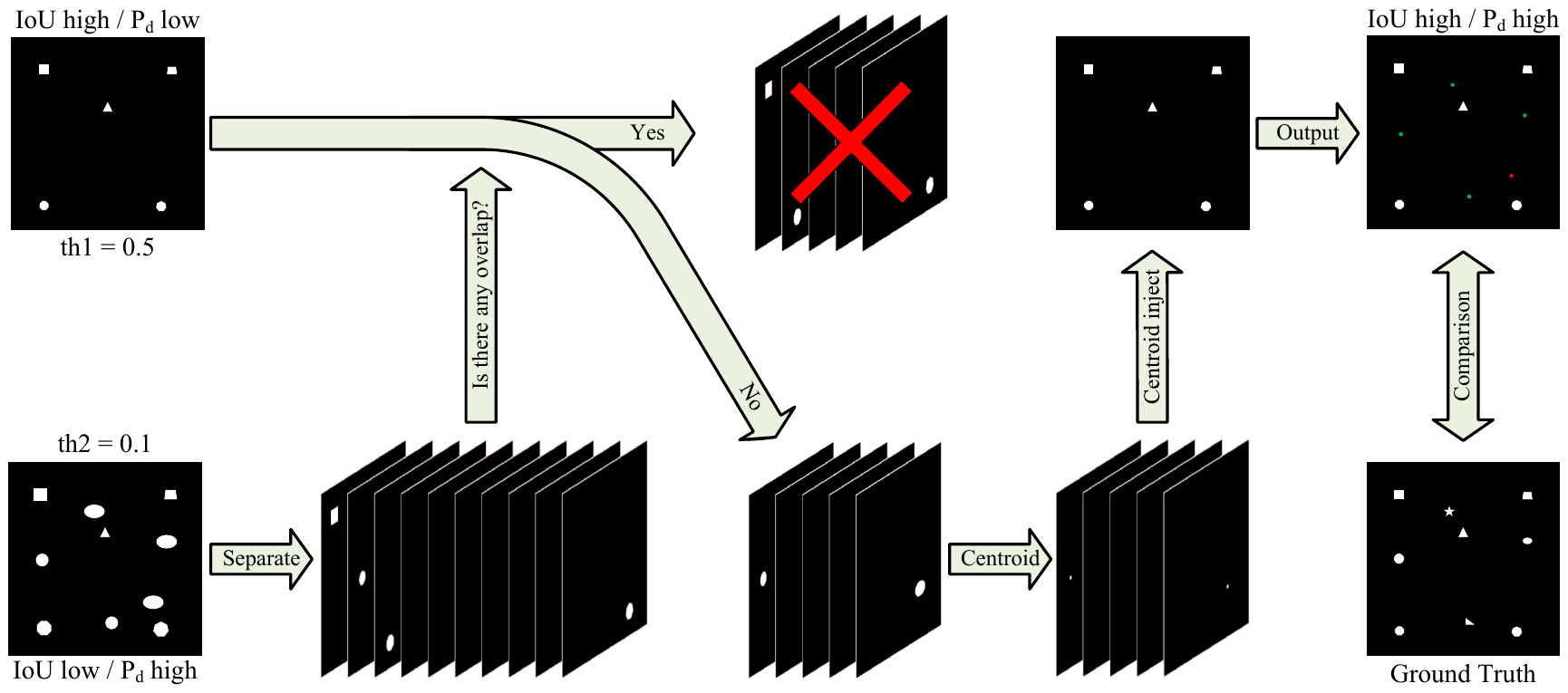}
	\vspace{-13pt}
	\caption{\hla{Adjustable sensitivity strategy. Each connected component in the figure denotes an infrared small target.} \textcolor{green}{\textbf{\hla{Green}}} \hla{denotes correctly included weak targets, while} \textcolor{lightred}{\textbf{\hla{red}}} \hla{denotes incorrectly included weak targets.}}
	\label{fig:fig_method_as}
	\vspace{-5pt}
\end{figure}

The specific process of the AS strategy is shown in \Cref{fig:fig_method_as}. First, we separate the segmentation result ${M_{th2}}$ of $th2$ into multiple binary images $M_{th2}^s,\;s = 1,2, \ldots n$ with a single target region according to a single connected region. Secondly, these binary images $M_{th2}^s,\;s = 1,2, \ldots n$ are compared with the segmentation results ${M_{th1}}$ of $th1$ in turn to determine whether there are overlapping regions. For $M_{th2}^s$ with overlapping regions, we lose it. For the same target region, the fineness in ${M_{th1}}$ is better than that in ${M_{th2}}$. Thirdly, we extract the centroid of $M_{th2}^s$ where there is no overlapping region. Finally, the extracted centroid points are injected into ${M_{th1}}$ to generate the final segmentation result map. The highlighted weak target region is added to ${M_{th1}}$ in the form of centroid points without affecting the fine segmentation of the strong target region in ${M_{th1}}$. It is worth noting that this strategy exhibits good generality, as it can be seamlessly integrated into existing networks as a post-processing step without requiring any modifications to the original structure.

\section{Experiment}
\label{sec:experiment}
\subsection{Dataset}
\begin{figure*}[t]
        \vspace{0pt}
	\centering
	\includegraphics[width=\textwidth]{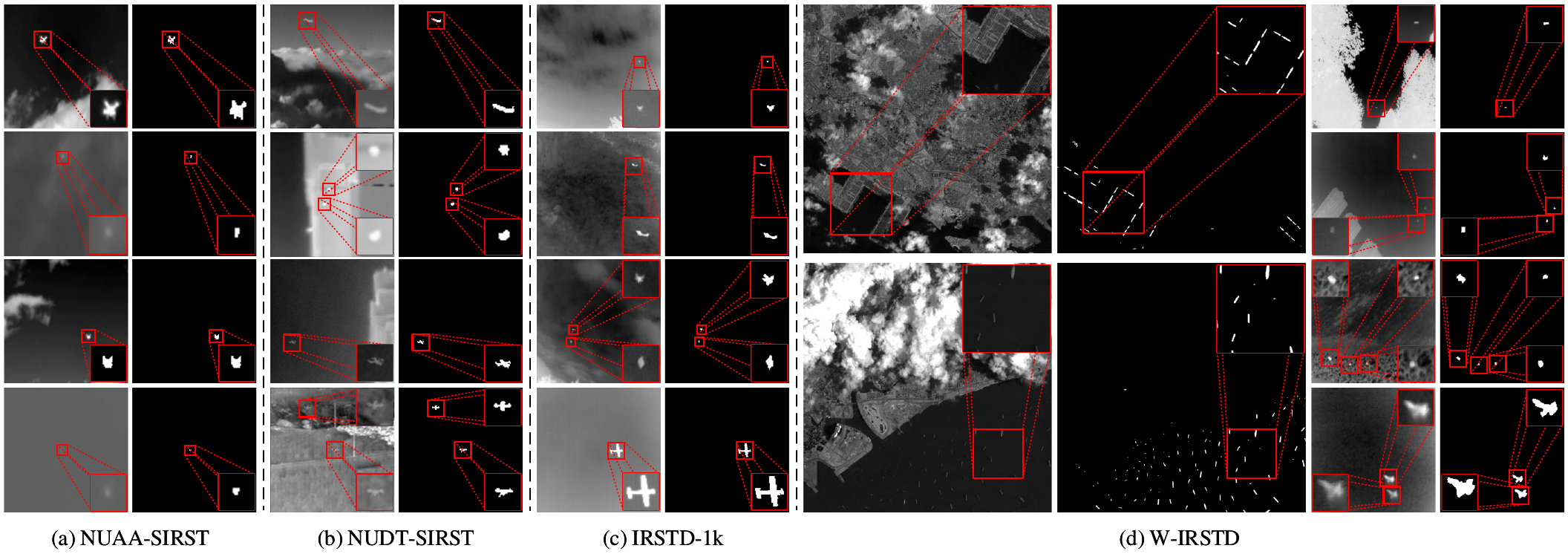}
	\vspace{-16pt}
	\caption{\hla{Some samples from multiple datasets.}}
	\label{fig:fig_ex_dataset_intro}
	\vspace{-3pt}
\end{figure*}

Four SIRST datasets (W-IRSTD dataset~\cite{li2024icpr}, NUDT-SIRST~\cite{li2022dense}, NUAA-SIRST~\cite{dai2021asymmetric}, and IRSTD-1k~\cite{zhang2022isnet}) are used for \hla{the} experiments. These datasets are introduced as follows:

\textbf{\textit{1) W-IRSTD dataset:}} The dataset contains 9,000 images, which mainly come from seven public datasets: SIRST-V2~\cite{dai2023one}, IRSTD-1k~\cite{zhang2022isnet}, IRDST~\cite{sun2023receptive}, NUDT-SIRST~\cite{li2022dense}, NUDT-SIRST-sea~\cite{wu2023mtu}, NUDT-MIRSDT~\cite{li2023direction} and Anti-UAV410~\cite{huang2023anti}. \Cref{fig:fig_ex_dataset_intro}(d) shows some samples from the W-IRSTD dataset. The dataset covers multiple observation perspectives, including land-based, air-based, and space-based. In addition, it includes diverse target types such as extended targets, spot targets, and point targets, spans multiple spectral bands including short-wave infrared, long-wave infrared, and near-infrared, and provides various resolutions such as 256×256, 512×512, and 1024×1024. In the experiment, 7200 samples are used as the training set and 1800 samples are used as the test set.

\textbf{\textit{2) NUAA-SIRST dataset:}} The real dataset consists of 427 infrared images captured from various scenes, with approximately 10\% of the images containing multiple targets. In the experiment, we divide the dataset into \hla{a training set and a test set at a ratio of 213:214.} \Cref{fig:fig_ex_dataset_intro}(a) shows some samples from the NUAA-SIRST dataset.

\textbf{\textit{3) NUDT-SIRST dataset:}} The synthetic dataset covers five background types: city, field, highlight, ocean and cloud. \hla{Approximately} 37\% of the images contain two or more targets, and \hla{approximately} 32\% of the targets are distributed outside the top 10\% of the image brightness values. In the experiment, we divide the dataset into \hla{a training set and a test set at a ratio of 663:664.} \Cref{fig:fig_ex_dataset_intro}(b) shows some samples from the NUDT-SIRST dataset.

\textbf{\textit{4) IRSTD-1k dataset:}} The real dataset consists of 1001 infrared images with a resolution of 512×512 pixels. It contains \hla{various types of small targets}, including drones, creatures, ships, and vehicles, with diverse spatial distributions. Meanwhile, the dataset covers a wide range of scenes, such as \hla{the} sky, ocean, and land. In the experiments, we divide the dataset into \hla{a training set and a test set at a ratio of 801:200.} \Cref{fig:fig_ex_dataset_intro}(c) shows some samples from the IRSTD-1k dataset.

\subsection{Experimental settings}
\textbf{\textit{1) Experimental environment and parameter settings.}} The experimental environment is an Ubuntu 18.04 operating system, and the GPU is an RTX 3090 24 GB. \hla{The number of training epochs} is set to 300 and the learning rate is set to $5e^{-4}$.

\begin{figure}[t]
        \vspace{0pt}
	\centering
	\includegraphics[width=\columnwidth]{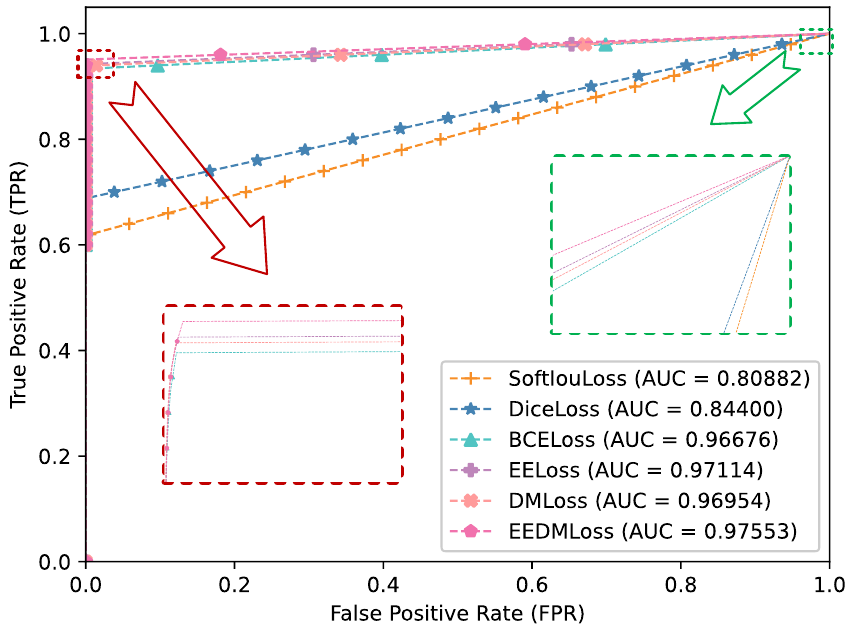}
	\vspace{-18pt}
	\caption{ROC curves of models trained with different loss functions.}
	\label{fig:fig_ex_loss_roc}
	\vspace{-5pt}
\end{figure}

\begin{table}[t]
\begin{minipage}{\columnwidth}
\centering
\caption{Performance comparison of different losses on the W-IRSTD dataset.}
\vspace{-5pt}
\label{tab:tab01}
\renewcommand{\arraystretch}{1.1}
\setlength{\tabcolsep}{3mm}{
\resizebox{\columnwidth}{!}{
\begin{tabular}{ccccc}
\hline
Scheme                & $IoU$ ↑           & $P_d$ ↑            & $F_a$ (×$10^{-6}$) ↓   & \textbf{\textit{Score}} ↑         \\ \hline
SoftIoU Loss \textsuperscript{\cite{rahman2016optimizing}} & 44.08          & 53.69          & 12.39         & 48.89          \\
Dice Loss \textsuperscript{\cite{li2019dice}}    & 50.15          & 60.09          & 9.09          & 55.12          \\
BCE Loss \textsuperscript{\cite{li2024rediscovering}}     & 55.44          & 59.35          & \textbf{7.02} & 57.40          \\ \hline
EE Loss (ours)        & 56.26          & 60.46          & 8.51          & 58.36          \\
DM Loss (ours)        & 56.12          & 62.45          & 7.83          & 59.29          \\ \hline
\rowcolor[HTML]{E7E6E6} 
EEDM Loss (ours)      & \textbf{57.30} & \textbf{63.34} & 14.25         & \textbf{60.32} \\ \hline
\end{tabular}
}}
\vspace{13pt}
\end{minipage}

\begin{minipage}{\columnwidth}
\centering
\caption{Performance comparison of different scales on the W-IRSTD dataset.}
\vspace{-5pt}
\label{tab:tab02}
\renewcommand{\arraystretch}{1.0}
\setlength{\tabcolsep}{4mm}{
\resizebox{\columnwidth}{!}{
\begin{tabular}{ccccc}
\hline
Scheme & $IoU$ ↑          & $P_d$ ↑           & $F_a$ (×$10^{-6}$) ↓   & \textbf{\textit{Score}} ↑        \\ \hline
256    & 43.78          & 48.75          & 13.79          & 46.27          \\
384    & 53.79          & 64.89          & \textbf{11.19} & 59.34          \\
512    & 57.30          & 63.34          & 14.25          & 60.32          \\
640    & 59.65          & 72.05          & 11.55          & 65.85          \\
768    & 60.67          & 70.90          & 13.53          & 65.79          \\
896    & \textbf{61.89} & 71.95          & 12.32          & 66.92          \\
1024   & 60.13          & \textbf{76.17} & 12.78          & \textbf{68.15} \\ \hline
\end{tabular}
}}
\end{minipage}
\end{table}

\textbf{\textit{2) Evaluation metrics.}} This study uses pixel-level evaluation metrics (intersection over union, $IoU$), target-level evaluation metrics (detection rate $P_d$ and false alarm rate $F_a$) and comprehensive evaluation metrics \textbf{\textit{Score}} to evaluate the performance of the model. The metric \textbf{\textit{Score}} is calculated by the $IoU$ and $P_d$ under a certain $F_a$ constraint. The \textbf{\textit{Score}} can be used to comprehensively evaluate the region segmentation accuracy and target detection accuracy of the model:
\begin{gather}
IoU = \frac{{\sum\nolimits_i^N {TP[i]} }}{{\sum\nolimits_i^N {T[i] + p[i] - TP[i]} }} \\[3pt]
{P_d} = \frac{{{T_{correct}}}}{{{T_{All}}}}  \\[3pt]
{F_a} = \frac{{{P_{false}}}}{{{P_{all}}}}   \\[3pt]
Score = \alpha  \times IoU + (1 - \alpha ) \times {P_d}   
\end{gather}
where $N$ denotes the number of images in the test set. $TP$ denotes the small target area that the model correctly predicts. $T$ denotes the small target area in the true label. $P$ denotes the small target area predicted by the model. ${T_{correct}}$ denotes the number of correctly predicted targets. ${T_{All}}$ denotes the true target number. ${P_{false}}$ denotes the number of incorrectly predicted target pixels. ${P_{all}}$ denotes the total number of pixels in the image. $\alpha$ is set to 0.5. In addition, $F_a$ has a constraint effect. When $F_a$ is less than $10^{-4}$, the final result is considered valid.

\subsection{Effect verification of the EEDM loss}
To further verify the effect of the proposed EEDM loss, we conduct \hla{extensive experiments, including comparisons with other common losses~\cite{li2024rediscovering,rahman2016optimizing,li2019dice} and an evaluation of hyperparameter stability.} The experiments are conducted on the W-IRSTD dataset. MSDA-Net is selected as the basic model.

\textbf{\textit{1) Comparison with other loss functions:}} To evaluate the effectiveness of the proposed \hla{EEDM loss, we conduct comparative experiments against commonly used loss functions, including binary cross-entropy (BCE) loss~\cite{li2024rediscovering}, soft intersection over union (SoftIoU) loss~\cite{rahman2016optimizing} and Dice loss~\cite{li2019dice}. To comprehensively evaluate the effectiveness of the proposed EEDM loss, we further introduce two ablation variants for comparison: the edge enhancement (EE) loss, which contains only the edge enhancement part, and the difficult mining (DM) loss}, which contains only the difficult pixel mining part. This design enables a quantitative analysis of the respective contributions of edge enhancement and difficult sample mining to the overall detection performance.

\begin{figure}[t]
        \vspace{0pt}
	\centering
	\includegraphics[width=\columnwidth]{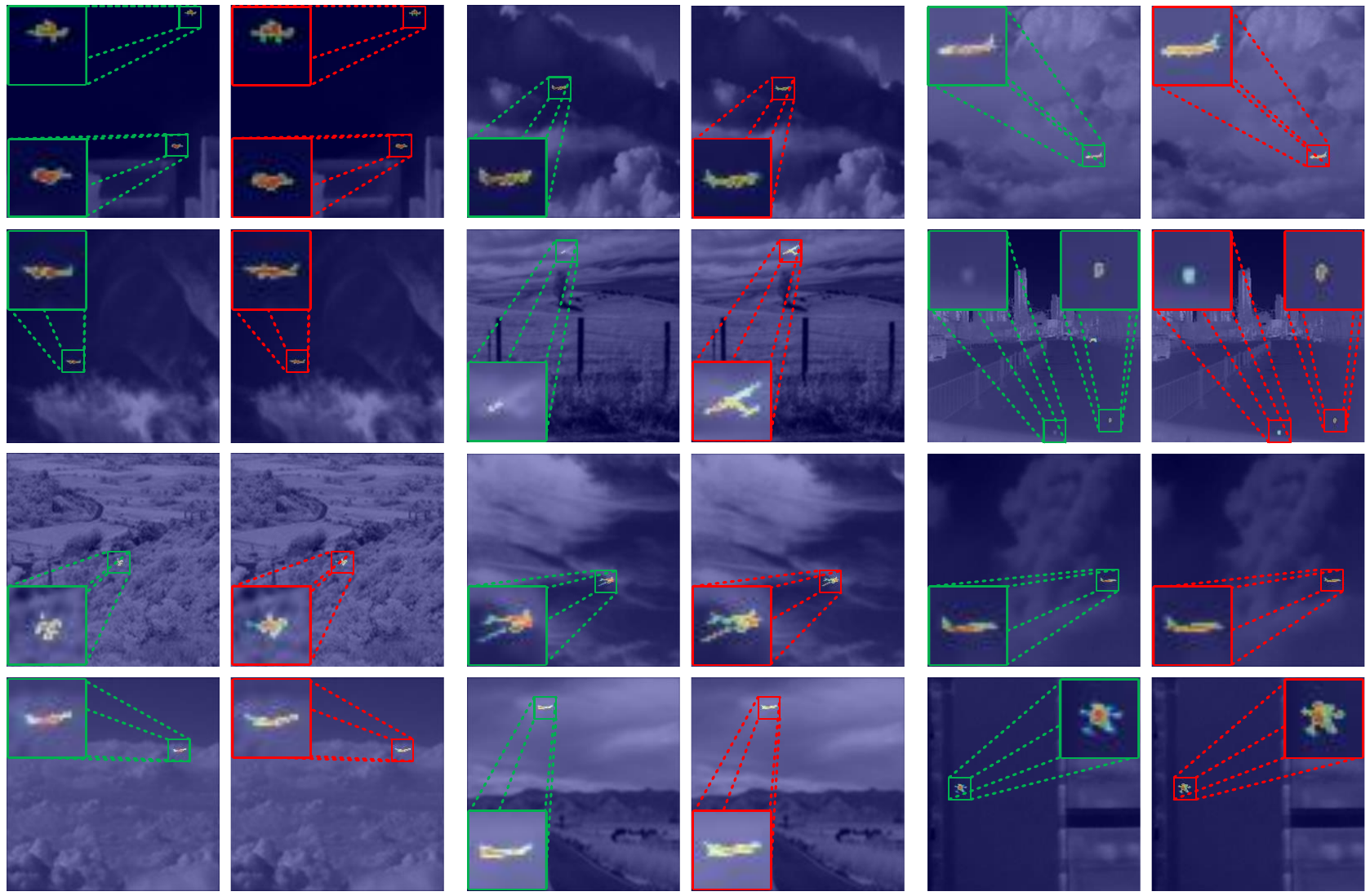}
	\vspace{-18pt}
	\caption{\hla{Grad-CAM visualization comparison of detection results with SoftIoU Loss and EEDM Loss. In each group of images, the ones with green borders denote detection results using SoftIoU Loss, while the ones with red borders denote detection results using EEDM Loss. Local zoom has been applied to emphasize the differences in key areas.}}
	\label{fig:fig_ex_loss_grad_cam}
	\vspace{-5pt}
\end{figure}

From \Cref{tab:tab01}, compared with other losses, the proposed EEDM loss results in more obvious improvements in both \hla{the} $IoU$ and $P_d$, verifying its superiority in infrared small target detection tasks. In contrast, the widely used \hla{SoftIoU loss} exhibits relatively inferior performance on the complex and diverse W-IRSTD dataset. At the same time, compared with \hla{BCE loss, EE loss} achieves improvements of 1.48\% in \hla{the} $IoU$ (from 55.44 to 56.26), 1.87\% in \hla{the} $P_d$ (from 59.35 to 60.46), and 1.67\% in \hla{the} \textbf{\textit{Score}} (from 57.40 to 58.36), indicating that incorporating edge features during training can effectively enhance detection performance. In addition, compared with \hla{BCE loss, DM loss} achieves improvements of 1.23\% in \hla{the} $IoU$ (from 55.44 to 56.12), 5.2\% in \hla{the} $P_d$ (from 59.35 to 62.45), and 3.19\% in \hla{the} \textbf{\textit{Score}} (from 57.40 to 59.29), indicating the effectiveness of difficult pixel mining in this task. From \Cref{fig:fig_ex_loss_roc}, the ROC curve of the model trained with \hla{EEDM loss} consistently stays above the others, achieving the highest AUC, which indicates superior performance across all thresholds compared with other loss functions. \hla{Furthermore, to intuitively demonstrate the performance of EEDM loss, we select edge-rich infrared small target samples and visualize the detection results using Grad-CAM \cite{selvaraju2017grad}. Specifically, we compare the detection results obtained with the commonly used SoftIoU loss and the proposed EEDM loss. From \Cref{fig:fig_ex_loss_grad_cam}, compared with the SoftIoU loss, the EEDM loss yields more precise and sensitive responses to target boundaries and fine-grained structures.}

\begin{table}[t]
\begin{minipage}{\columnwidth}
\centering
\caption{Comparison of the fusion performance of multi-scale models on the W-IRSTD dataset.}
\vspace{-5pt}
\label{tab:tab03}
\renewcommand{\arraystretch}{1.0}
\setlength{\tabcolsep}{1.5mm}{
\resizebox{\columnwidth}{!}{
\begin{tabular}{ccccc}
\hline
Scheme                       & $IoU$ ↑          & $P_d$ ↑           & $F_a$ (×$10^{-6}$) ↓  & \textbf{\textit{Score}} ↑        \\ \hline
896                          & 61.89          & 71.95          & 12.32         & 66.92          \\
896+768                      & 63.12          & 70.85          & 9.32          & 66.99          \\
896+768+1024                 & \textbf{64.49} & \textbf{72.58} & 9.17          & \textbf{68.54} \\
896+768+1024+640             & 64.11          & 72.42          & 7.96          & 68.27          \\
896+768+1024+640+512         & 63.60          & 70.43          & 7.25          & 67.02          \\
896+768+1024+640+512+384     & 62.83          & 68.72          & 6.75          & 65.78          \\
896+768+1024+640+512+384+256 & 61.44          & 65.94          & \textbf{6.17} & 63.69          \\ \hline
\end{tabular}
}}
\vspace{10pt}
\end{minipage}

\begin{minipage}{\columnwidth}
\centering
\caption{Performance comparison of adjusting the threshold $th1$ on the W-IRSTD dataset.}
\vspace{-5pt}
\label{tab:tab04}
\renewcommand{\arraystretch}{1.0}
\setlength{\tabcolsep}{1.5mm}{
\resizebox{\columnwidth}{!}{
\begin{tabular}{cccccc}
\hline
Scheme                      & $th1$          & $IoU$ ↑          & $P_d$ ↑           & $F_a$ (×$10^{-6}$) ↓                  & \textbf{\textit{Score}} ↑        \\ \hline
896+768+1024 (0.5)          & 0.5          & 64.49          & 72.58          & \textbf{9.17}                 & 68.54          \\
896+768+1024 (0.45)         & 0.45         & 65.53          & 74.86          & 10.51                         & 70.20          \\
896+768+1024 (0.4)          & 0.4          & \textbf{65.91} & 76.44          & 13.03                         & 71.18          \\
896+768+1024 (0.35)         & 0.35         & 65.58          & 78.56          & 15.67                         & 72.07          \\
\textbf{896+768+1024 (0.3)} & \textbf{0.3} & 64.22          & 80.29          & 20.57                         & \textbf{72.26} \\
896+768+1024 (0.25)         & 0.25         & 61.83          & 82.34          & 26.85                         & 72.09          \\
896+768+1024 (0.2)          & 0.2          & 58.34          & 83.91          & 35.93                         & 71.13          \\
896+768+1024 (0.15)         & 0.15         & 53.43          & 85.28          & 50.52                         & 69.36          \\
896+768+1024 (0.1)          & 0.1          & 46.27          & 86.43          & 76.12                         & 66.35          \\
896+768+1024 (0.05)         & 0.05         & 35.15          & \textbf{87.72} & {\color[HTML]{FF0000} 138.46} & --              \\ \hline
\end{tabular}
}}
\vspace{10pt}
\end{minipage}

\begin{minipage}{\columnwidth}
\centering
\caption{Performance comparison of adjusting the threshold $th2$ on the W-IRSTD dataset.}
\vspace{-5pt}
\label{tab:tab05}
\renewcommand{\arraystretch}{1.0}
\setlength{\tabcolsep}{2mm}{
\resizebox{\columnwidth}{!}{
\begin{tabular}{ccccccc}
\hline
Scheme                          & $th1$                  & $th2$          & $IoU$            & $P_d$             & $F_a$ (×$10^{-6}$)     & \textbf{\textit{Score}}         \\ \hline
896+768+1024 (0.3)              & \multirow{7}{*}{0.3} & -            & \textbf{64.22} & 80.29          & \textbf{20.57} & 72.26          \\
896+768+1024 (0.3+0.5)          &                      & 0.5          & 63.96          & 81.61          & 21.26          & 72.79          \\
896+768+1024 (0.3+0.4)          &                      & 0.4          & 63.78          & 83.02          & 21.76          & 73.40          \\
896+768+1024 (0.3+0.3)          &                      & 0.3          & 63.45          & 85.28          & 22.66          & 74.37          \\
896+768+1024 (0.3+0.2)          &                      & 0.2          & 62.81          & 87.51          & 24.29          & 75.16          \\
\textbf{896+768+1024 (0.3+0.1)} &                      & \textbf{0.1} & 61.42          & 89.98          & 28.11          & \textbf{75.70} \\
896+768+1024 (0.3+0.05)         &                      & 0.05         & 59.44          & \textbf{91.05} & 33.05          & 75.25          \\ \hline 
\end{tabular}
}}
\end{minipage}
\end{table}

\textbf{\textit{2) Validation of hyperparameter stability:}} To comprehensively evaluate the performance of the proposed EEDM loss, we conduct \hla{detailed} experiments on two key hyperparameters (the edge weight coefficient $w$ and the difficult pixel mining ratio $p$) to investigate the model's sensitivity and robustness to different hyperparameter settings. Specifically, the edge weight coefficient $w$ is set to {1, 3, 4, 5, 7}, and the difficult pixel mining ratio $p$ is set to {0.1, 0.3, 0.5, 0.7, 0.9}.

\begin{figure*}[t]
        \vspace{0pt}
	\centering
	\includegraphics[width=\textwidth]{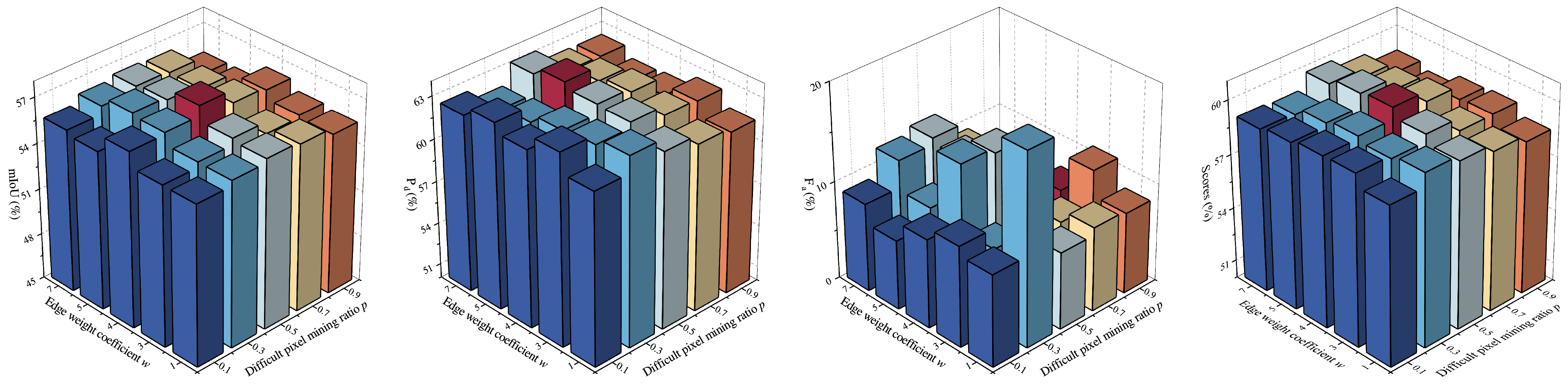}
	\vspace{-10pt}
	\caption{Performance analysis of the hyperparameters $w$ and $p$ on the W-IRSTD dataset. From left to right, the figures show the model performance in terms of the $IoU$, $P_d$, $F_a$, and \textbf{\textit{Score}} under different combinations of $(w, p)$. \textcolor{red}{\textbf{Red}} indicates the optimal value of each metric.}
	\label{fig:fig_ex_hyper_p}
	\vspace{-5pt}
\end{figure*}

From \Cref{fig:fig_ex_hyper_p}, the proposed EEDM \hla{loss} can maintain relatively stable performance across different hyperparameter configurations. This indicates that the loss function is robust to the hyperparameters $w$ and $p$, as it consistently enhances model performance within a reasonable range of these hyperparameters. Based on the experimental results, the metric \textbf{\textit{Score}} reaches the optimal value when $w$ is 4 and $p$ is 0.5. \hla{Specifically,} when $w$ is too small, the model pays insufficient attention to edge information, leading to decreases in \hla{the} $IoU$ and $P_d$. \hla{Conversely, when} $w$ is too large, the model will overly rely on edge features, thereby neglecting semantic and other critical features, which negatively impacts the overall detection performance. It is worth noting that when $w$ is extremely reduced to 1, the EEDM \hla{loss} degenerates into the DM \hla{loss}. \hla{Meanwhile, when $p$ is too small,} the network tends to excessively focus on difficult pixels during training, which may hinder its generalization ability and lead to degraded performance. Conversely, when $p$ is too large, the mining of difficult pixels will be insufficient, thereby weakening the model's adaptability in complex scenarios. It is worth noting that when $p$ increases to 1, the EEDM loss degenerates into the EE loss.

\subsection{Effect verification of the multi-scale fusion}
To evaluate the effectiveness of the multi-scale fusion strategy, we conduct extensive experiments on the W-IRSTD dataset, using MSDA-Net as the basic model and EEDM loss as the loss function. Specifically, we resize the original images to various resolutions, including 256×256, 384×384, 512×512, 640×640, 768×768, 896×896, and 1024×1024, to systematically explore the impact of \hla{the} input scale on model performance. \hla{On this basis,} we take the best-performing single-scale model as the foundation and progressively integrate suboptimal models to further evaluate the effectiveness of the multi-scale fusion strategy. For clarity, each model is named based on its corresponding input resolution. For example, the model trained with 256×256 images is denoted as the 256 model.

From \Cref{tab:tab02}, the 1024 model achieves the best performance. Compared with the worst performing 256 model, the 1024 model improves the $IoU$, $P_d$, and \textbf{\textit{Score}} by 37.35\% (from 43.78 to 60.13), 56.25\% (from 48.75 to 76.17) and 47.29\% (from 46.27 to 68.15), respectively. The analysis reveals that when images from the W-IRSTD dataset are downsampled to smaller resolutions, such as 256×256, the model performance drops significantly. This degradation is primarily attributed to the loss of pixel-level information during the downsampling process, which causes small targets to shrink or even vanish entirely, making it difficult for the network to extract effective features. In addition, considering that the proposed AS strategy will be used later, it can effectively improve the detection rate of the model. Therefore, to ensure segmentation accuracy, we choose the 896 model with the best $IoU$ metric as the basis model, and fuse the suboptimal scale models in the $IoU$ metric one by one. The experimental results are shown in \Cref{tab:tab03}.

\begin{figure}[!t]
        \vspace{0pt}
	\centering
	\includegraphics[width=1\columnwidth]{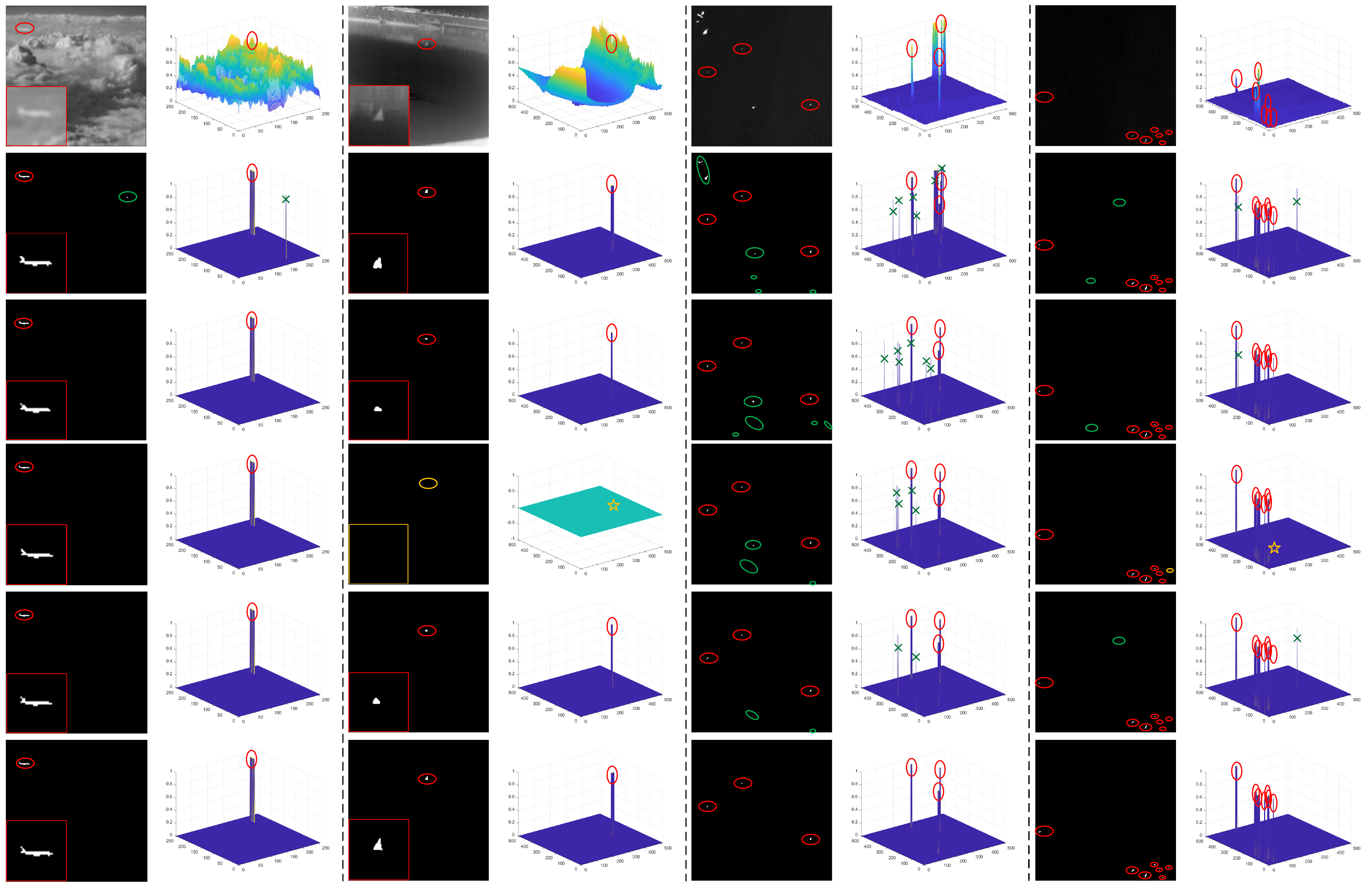}
	\vspace{-16pt}
	\caption{Detection results of the fusion model and its corresponding single model. From top to bottom, each row sequentially displays the original image, detection results from the 768 model, 896 model, 1024 model, and the fused model, followed by the corresponding ground truth. In the visualizations,  \textcolor{lightred}{\textbf{red}} denotes the correct detections, \textcolor{green}{\textbf{green}} denotes the false detections, and \textcolor{yellow}{\textbf{yellow}} denotes missed detections.}
	\label{fig:fig_ex_3d_vis}
	\vspace{3pt}
\end{figure}

From \Cref{tab:tab03}, the fusion of the 896, 768, and 1024 models yields the most significant performance improvement. Specifically, compared with the 896 model, which has the best performance at a single scale, the fusion of the 896 model, the 768 model and the 1024 model can improve its $IoU$, $P_d$ and \textbf{\textit{Score}} by 4.20\% (from 61.89 to 64.49), 0.88\% (from 71.95 to 72.58) and 2.42\% (from 66.92 to 68.54), respectively. At the same time, we find that when the performance of the fused scale models degrades severely, it will lead to the degradation of the performance of the final model. In addition, the multi-scale fusion model will perform better on $F_a$, which shows that the multi-scale fusion strategy is conducive to reducing the missed detections and further improving the reliability of the model. To more intuitively demonstrate the effect of multi-scale model fusion, we show the prediction results of the 896 model, 768 model, 1024 model and their fusion model in \Cref{fig:fig_ex_3d_vis}. It is worth noting that although multi-scale model fusion \hla{achieves} significant improvements in detection performance, it inevitably increases \hla{the} inference time. Therefore, a reasonable trade-off between accuracy and efficiency should be made based on specific requirements in practical applications to achieve optimal overall performance.

\subsection{Effect verification of the AS strategy}
\hla{To validate the AS strategy}, we adjust the thresholds $th1$ and $th2$ on the optimal multi-model fusion scheme (“896+768+1024”). \Cref{tab:tab04} shows the experimental results obtained by adjusting the threshold $th1$ alone without adjusting the threshold $th2$. \Cref{tab:tab05} shows the experimental results obtained by adjusting the threshold $th2$ on the basis that $th1$ is set to the optimal value in \Cref{tab:tab04}.

From \Cref{tab:tab04}, when the threshold $th1$ is set to 0.3, the model achieves the best overall performance. Compared \hla{with} the conventional setting of $th1$ at 0.5, adjusting $th1$ to 0.3 leads to 5.43\% (from 68.54 to 72.26) improvement in the comprehensive metric \textbf{\textit{Score}}. The key to this improvement lies in the model's ability to significantly \hla{increase} the detection rate $P_d$ without notably affecting \hla{the} segmentation accuracy. Specifically, $P_d$ increases by 10.62\% (from 72.58 to 80.29). Notably, when the threshold $th1$ is reduced to 0.05, $F_a$ will surge and exceed the constraint range.

From \Cref{tab:tab05}, the introduction of $th2$ significantly improves the detection rate of the network while maintaining segmentation accuracy for infrared small targets and effectively controlling the false alarm rate. \hla{On the W-IRSTD dataset, the model achieves optimal performance under the multi-scale fusion scheme “896+768+1024” when $th1$ is set to 0.3 and $th2$ is set to 0.1. Specifically, compared with using only $th1$ set to 0.3, the further introduction of $th2$ set to 0.1 leads to a 4.76\% (from 72.26 to 75.70) improvement in \textbf{\textit{Score}} metric.}

\begin{figure}[!t]
        \vspace{0pt}
	\centering
	\includegraphics[width=1\columnwidth]{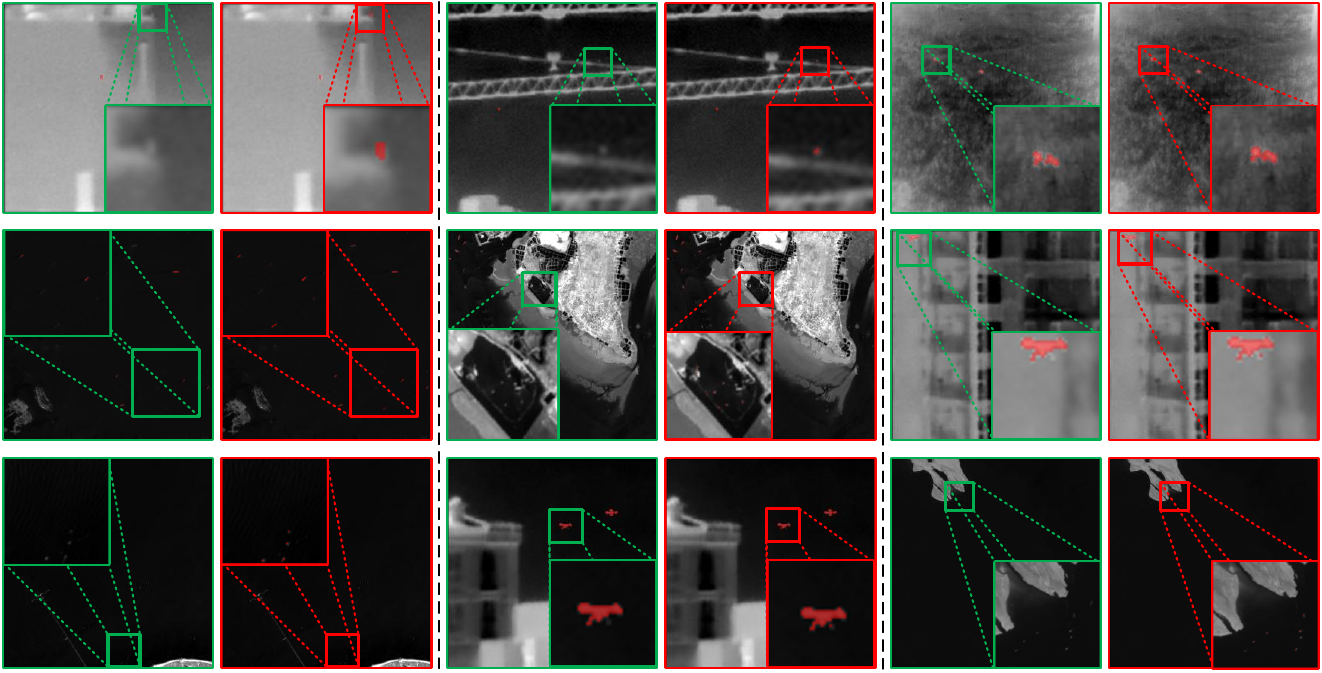}
	\vspace{-16pt}
	\caption{\hla{2D comparison of the effects of the AS strategy. In each group of images, the ones with} \textcolor{green}{\textbf{\hla{green}}} \hla{borders represent the detection results without the AS strategy, while the ones with}  \textcolor{lightred}{\textbf{\hla{red}}} \hla{borders represent the detection results after applying the AS strategy. Local zoom has been applied to emphasize the differences in key areas.}}
	\label{fig:fig_ex_2d_vis}
	\vspace{-3pt}
\end{figure}

To more intuitively demonstrate the effectiveness of the AS strategy, we present some results as shown in \Cref{fig:fig_ex_2d_vis}. For extended targets, the appropriate application of the AS strategy helps refine the shape and details of the target. For point or spot targets, \hla{this} strategy helps improve detection rates and effectively capture weak target features.

\subsection{Validation of the FEST Framework on multiple datasets}
To validate the effectiveness and robustness of the proposed FEST framework, we integrate multiple advanced SIRST networks into the framework and conduct experiments on three public datasets. To ensure a fair comparison, all \hla{the} models are retrained under the same settings, with the number of training epochs uniformly set to 300 to reduce training costs. This setup differs from that used in our previous work. \hla{In addition, we explore the performance gain brought by using only the EEDM loss and AS strategy in the FEST framework,} with almost no additional resource consumption. This lightweight variant is referred to as the LW-FEST framework. The thresholds $th1$ and $th2$ in both the FEST and LW-FEST frameworks are set to 0.5 and 0.1, respectively. Notably, considering the characteristics of each dataset, we adopt differentiated configurations in the multi-scale fusion strategy: for NUAA-SIRST and IRSTD-1k, we fuse models with sizes of 512 and 768, whereas for NUDT-SIRST, we fuse models with sizes of 256 and 384.

\hla{From \Cref{tab:tab06}, embedding various SIRST networks into the lightweight LW-FEST framework can effectively improve the model performance with minimal additional resource consumption.} These results demonstrate that the LW-FEST framework exhibits strong robustness and generalization, and can be seamlessly integrated with various SIRST networks, effectively enhancing their feature extraction and representation abilities. Meanwhile, when networks with relatively poor performance (such as ACM) are embedded into the framework, the performance improvement is more significant. In addition, the application of the complete FEST framework can further improve the detection performance. \hla{Specifically, on the NUAA-SIRST dataset, compared with the original SIRST network, the SIRST network equipped with the FEST framework achieves a performance improvement of 0.46\% - 7.40\% in the evaluation metric \textbf{\textit{Score}}. On the NUDT-SIRST dataset, compared with the original SIRST network, the SIRST network equipped with the FEST framework achieves a performance improvement of 0.48\% - 6.34\% in the evaluation metric \textbf{\textit{Score}}. On the IRSTD-1k dataset, compared with the original SIRST network, the SIRST network equipped with the FEST framework achieves a performance improvement of 1.66\% - 7.26\% in the evaluation metric \textbf{\textit{Score}}.}

\begin{figure*}[!t]
\begin{minipage}{\textwidth}
        \vspace{0pt}
	\centering
	\includegraphics[width=\textwidth]{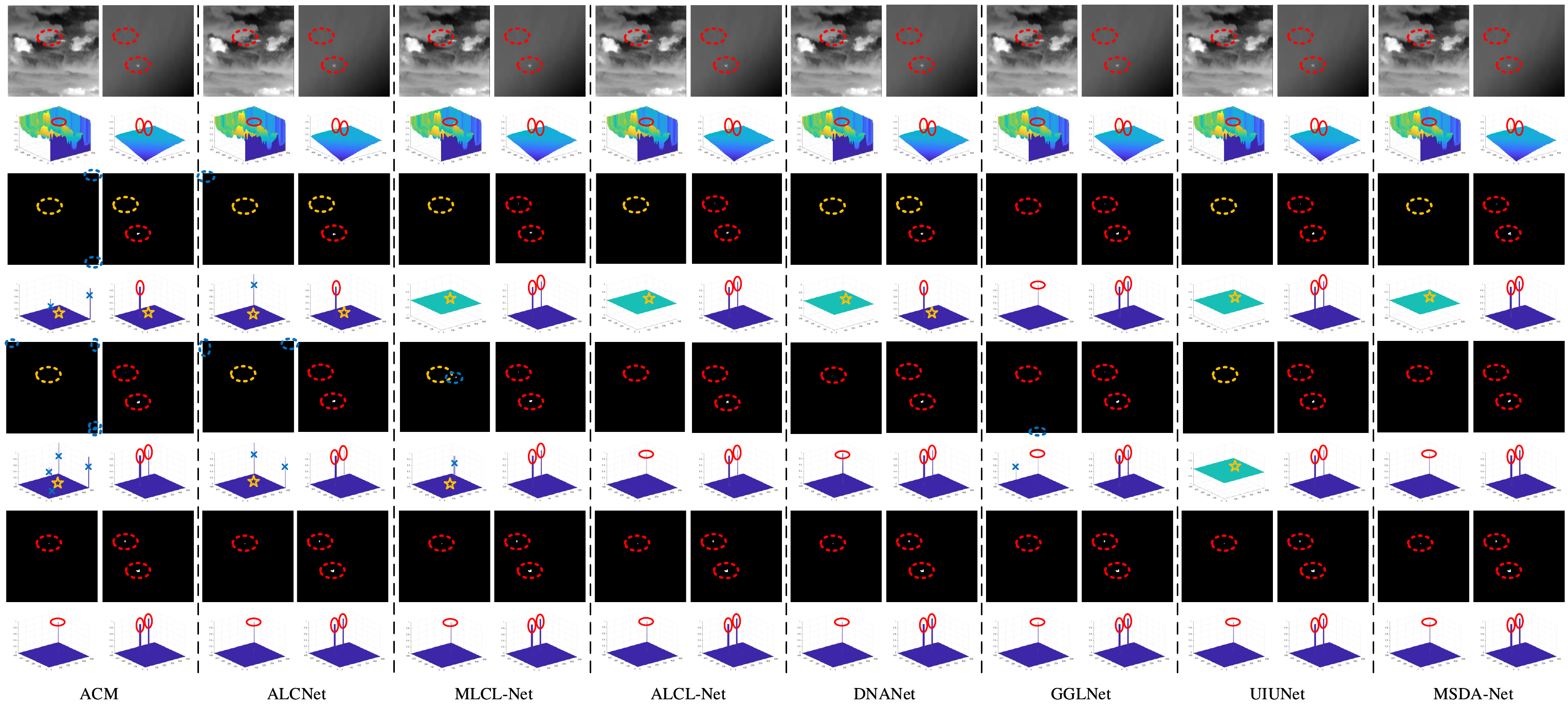}
	\vspace{-15pt}
	\caption{Visualization of several SIRST methods on the NUAA-SIRST dataset.  \textcolor{lightred}{\textbf{Red}} denotes correct detections, \textcolor{blue_mm}{\textbf{blue}} denotes false detections, and \textcolor{yellow}{\textbf{yellow}} denotes missed detections. Every two rows from top to bottom: Image, Original SIRST Network, SIRST Network equipped with FEST framework, True label.}
	\label{fig:fig_ex_vis_nuaa}
	\vspace{-5pt}
\end{minipage}
\end{figure*}

\begin{figure*}[!t]
\begin{minipage}{\textwidth}
        \vspace{0pt}
	\centering
	\includegraphics[width=\textwidth]{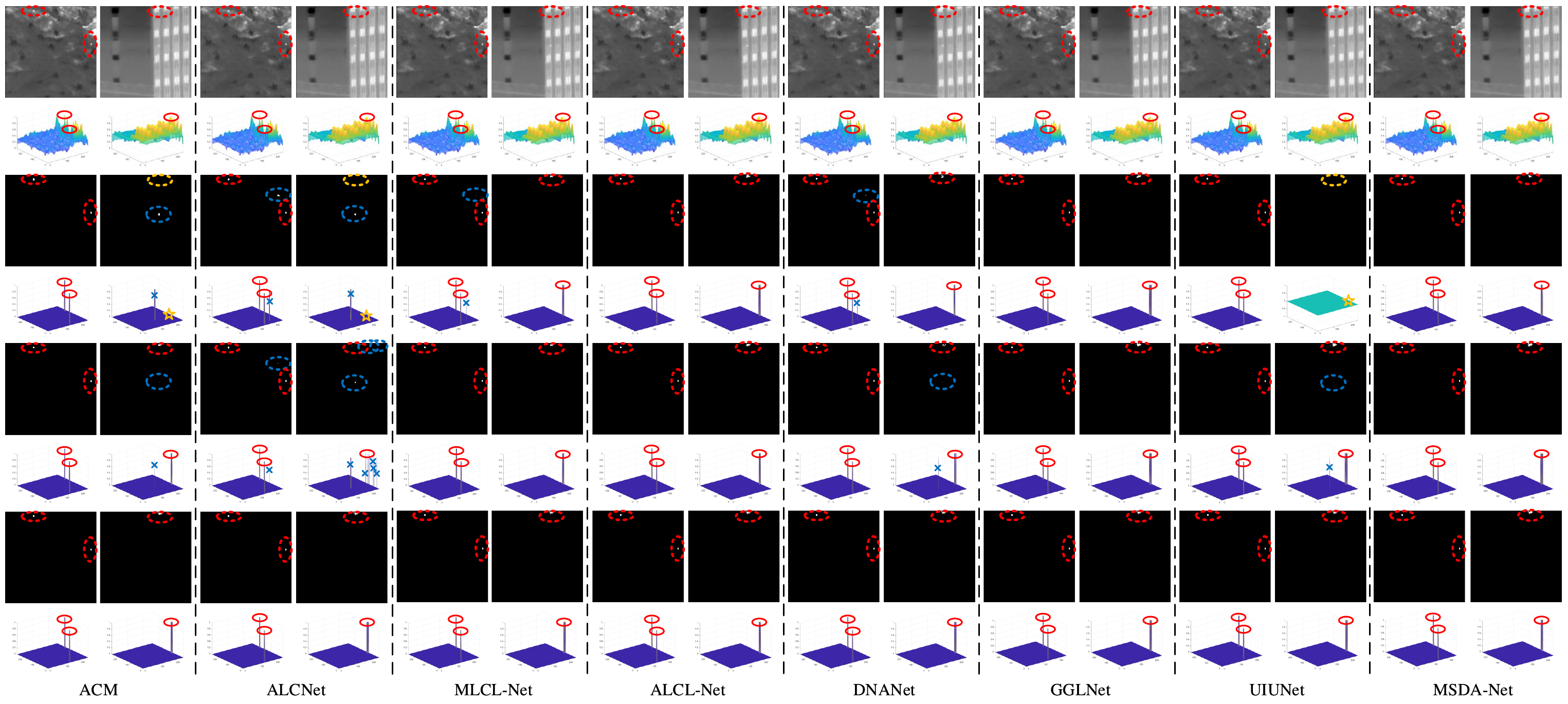}
	\vspace{-18pt}
	\caption{Visualization of several SIRST methods on the NUDT-SIRST dataset.  \textcolor{lightred}{\textbf{Red}} denotes correct detections, \textcolor{blue_mm}{\textbf{blue}} denotes false detections, and \textcolor{yellow}{\textbf{yellow}} denotes missed detections. Every two rows from top to bottom: Image, Original SIRST Network, SIRST Network equipped with FEST framework, True label.}
	\label{fig:fig_ex_vis_nudt}
	\vspace{-5pt}
\end{minipage}
\vspace{15pt}

\begin{minipage}{\textwidth}
        \vspace{0pt}
	\centering
	\includegraphics[width=\textwidth]{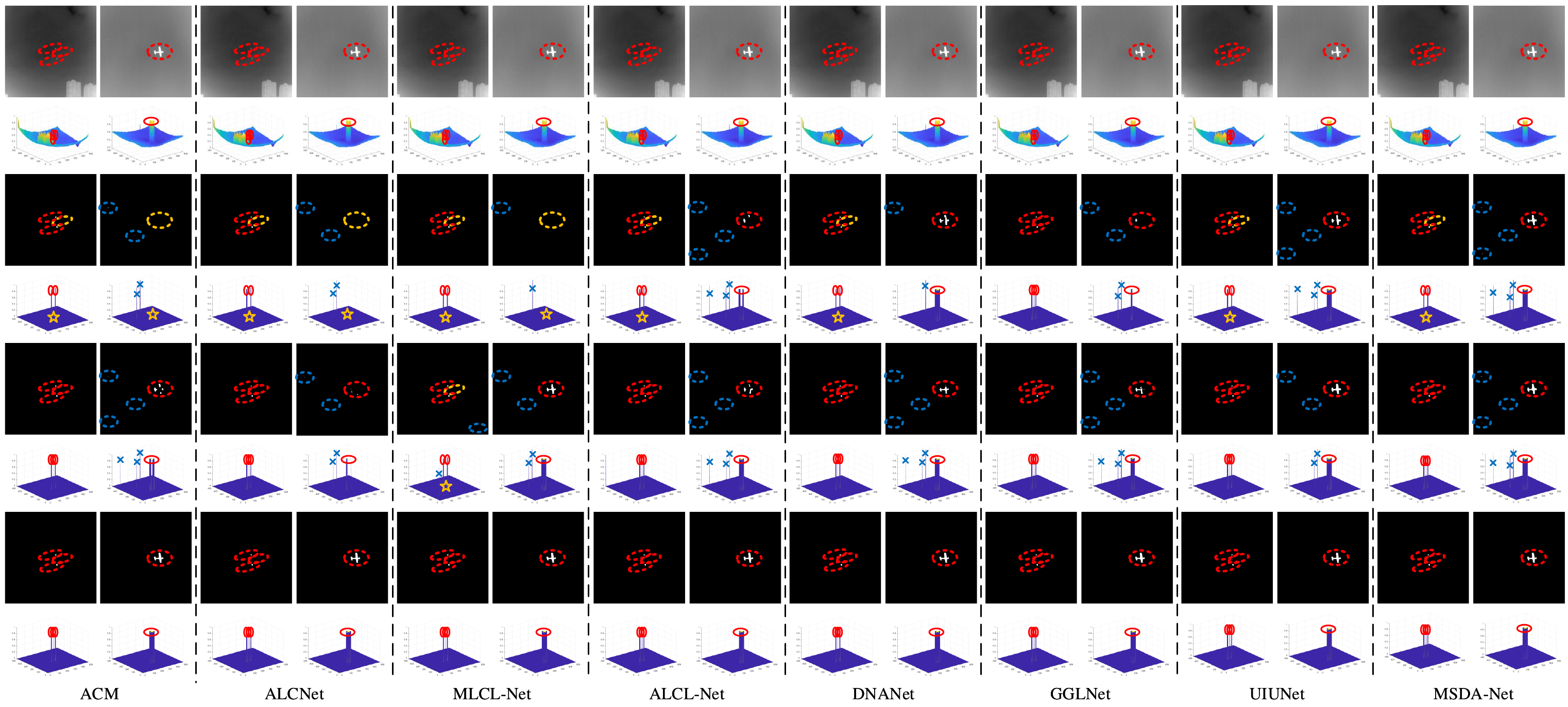}
	\vspace{-18pt}
	\caption{Visualization of several SIRST methods on the IRSTD-1k dataset.  \textcolor{lightred}{\textbf{Red}} denotes correct detections, \textcolor{blue_mm}{\textbf{blue}} denotes false detections, and \textcolor{yellow}{\textbf{yellow}} denotes missed detections. Every two rows from top to bottom: Image, Original SIRST Network, SIRST Network equipped with FEST framework, True label.}
	\label{fig:fig_ex_vis_irstd}
	\vspace{-5pt}
\end{minipage}
\end{figure*}

\hla{To further demonstrate the superiority of the proposed framework, we present visual comparisons of detection results on several samples across three datasets,} as shown in \Cref{fig:fig_ex_vis_nuaa}, \Cref{fig:fig_ex_vis_nudt}, and \Cref{fig:fig_ex_vis_irstd}. Equipping SIRST networks with the FEST framework leads to clearer target edges, more comprehensive detection results, and enhanced perception and localization accuracy for small targets in diverse detection scenarios.

\begin{table*}[!t]
\begin{minipage}{\textwidth}
\centering
\caption{\hla{$IoU$ (\%), $P_d$ (\%) and $F_a$ (×$10^{-6}$) values of different methods achieved on the separate NUAA-SIRST, NUDT-SIRST, and IRSTD-1k datasets. (213:214), (663:664) and (800:201) denote the division of training samples and test samples.}  \textcolor{lightred}{\textbf{\hla{Red}}} \hla{denotes that the $F_a$ exceeds the acceptable upper limit, making the results for this group invalid.} \textcolor{deepgreen}{\textbf{\hla{Green}}} \hla{denotes the best result within each group, where each group comprises a method’s Original, +LW-FEST, and +FEST variants.}}
\vspace{-5pt}
\label{tab:tab06}
\renewcommand{\arraystretch}{1.0}
\setlength{\tabcolsep}{2.5mm}{
\resizebox{\textwidth}{!}{
\begin{tabular}{c|c|cccc|cccc|cccc}
\hline
                           &                                  & \multicolumn{4}{c|}{NUAA-SIRST (213:214)}                                                                                                                                                                                                                     & \multicolumn{4}{c|}{NUDT-SIRST (663:664)}                                                                                                                                                                                                                     & \multicolumn{4}{c}{IRSTD-1k (800:201)}                                                                                                                                                                                                                        \\ \cline{3-14} 
\multirow{-2}{*}{Scheme}   & \multirow{-2}{*}{Description}    & $IoU$ ↑                                                          & $P_d$ ↑                                                          & $F_a$ ↓                                                          & $Score$ ↑                                                        & $IoU$ ↑                                                          & $P_d$ ↑                                                          & $F_a$ ↓                                                          & $Score$ ↑                                                        & $IoU$ ↑                                                          & $P_d$ ↑                                                          & $F_a$ ↓                                                          & $Score$ ↑                                                        \\ \hline
                           & Original                         & 62.24                                                         & 94.30                                                         & {\color[HTML]{FF0000} 111.84}                                 & -                                                             & 65.36                                                         & 96.08                                                         & 37.48                                                         & 80.72                                                         & 60.30                                                         & 89.56                                                         & {\color[HTML]{327640} \textbf{20.38}}                         & 74.93                                                         \\
                           & \cellcolor[HTML]{D7D7D7}+LW-FEST & \cellcolor[HTML]{D7D7D7}71.28                                 & \cellcolor[HTML]{D7D7D7}96.96                                 & \cellcolor[HTML]{D7D7D7}42.94                                 & \cellcolor[HTML]{D7D7D7}84.12                                 & \cellcolor[HTML]{D7D7D7}67.10                                 & \cellcolor[HTML]{D7D7D7}97.46                                 & \cellcolor[HTML]{D7D7D7}17.53                                 & \cellcolor[HTML]{D7D7D7}82.28                                 & \cellcolor[HTML]{D7D7D7}63.43                                 & \cellcolor[HTML]{D7D7D7}94.61                                 & \cellcolor[HTML]{D7D7D7}38.41                                 & \cellcolor[HTML]{D7D7D7}79.02                                 \\
\multirow{-3}{*}{ACM \textsuperscript{\cite{dai2021asymmetric}}}      & \cellcolor[HTML]{D7D7D7}+FEST    & \cellcolor[HTML]{D7D7D7}{\color[HTML]{327640} \textbf{73.57}} & \cellcolor[HTML]{D7D7D7}{\color[HTML]{327640} \textbf{97.72}} & \cellcolor[HTML]{D7D7D7}{\color[HTML]{327640} \textbf{34.78}} & \cellcolor[HTML]{D7D7D7}{\color[HTML]{327640} \textbf{85.65}} & \cellcolor[HTML]{D7D7D7}{\color[HTML]{327640} \textbf{73.16}} & \cellcolor[HTML]{D7D7D7}{\color[HTML]{327640} \textbf{98.52}} & \cellcolor[HTML]{D7D7D7}{\color[HTML]{327640} \textbf{9.86}}  & \cellcolor[HTML]{D7D7D7}{\color[HTML]{327640} \textbf{85.84}} & \cellcolor[HTML]{D7D7D7}{\color[HTML]{327640} \textbf{64.77}} & \cellcolor[HTML]{D7D7D7}{\color[HTML]{327640} \textbf{95.96}} & \cellcolor[HTML]{D7D7D7}33.54                                 & \cellcolor[HTML]{D7D7D7}{\color[HTML]{327640} \textbf{80.37}} \\ \hline
                           & Original                         & 66.96                                                         & 93.54                                                         & 67.83                                                         & 80.25                                                         & 70.84                                                         & 97.35                                                         & 24.40                                                         & 84.10                                                         & 62.93                                                         & 92.59                                                         & 29.87                                                         & 77.76                                                         \\
                           & \cellcolor[HTML]{D7D7D7}+LW-FEST & \cellcolor[HTML]{D7D7D7}70.19                                 & \cellcolor[HTML]{D7D7D7}97.34                                 & \cellcolor[HTML]{D7D7D7}38.42                                 & \cellcolor[HTML]{D7D7D7}83.77                                 & \cellcolor[HTML]{D7D7D7}71.03                                 & \cellcolor[HTML]{D7D7D7}97.88                                 & \cellcolor[HTML]{D7D7D7}20.63                                 & \cellcolor[HTML]{D7D7D7}84.46                                 & \cellcolor[HTML]{D7D7D7}64.26                                 & \cellcolor[HTML]{D7D7D7}95.62                                 & \cellcolor[HTML]{D7D7D7}15.68                                 & \cellcolor[HTML]{D7D7D7}79.94                                 \\
\multirow{-3}{*}{ALCNet \textsuperscript{\cite{dai2021attentional}}}   & \cellcolor[HTML]{D7D7D7}+FEST    & \cellcolor[HTML]{D7D7D7}{\color[HTML]{327640} \textbf{74.28}} & \cellcolor[HTML]{D7D7D7}{\color[HTML]{327640} \textbf{98.10}} & \cellcolor[HTML]{D7D7D7}{\color[HTML]{327640} \textbf{28.95}} & \cellcolor[HTML]{D7D7D7}{\color[HTML]{327640} \textbf{86.19}} & \cellcolor[HTML]{D7D7D7}{\color[HTML]{327640} \textbf{74.91}} & \cellcolor[HTML]{D7D7D7}{\color[HTML]{327640} \textbf{98.94}} & \cellcolor[HTML]{D7D7D7}{\color[HTML]{327640} \textbf{17.35}} & \cellcolor[HTML]{D7D7D7}{\color[HTML]{327640} \textbf{86.93}} & \cellcolor[HTML]{D7D7D7}{\color[HTML]{327640} \textbf{66.85}} & \cellcolor[HTML]{D7D7D7}{\color[HTML]{327640} \textbf{97.31}} & \cellcolor[HTML]{D7D7D7}{\color[HTML]{327640} \textbf{13.27}} & \cellcolor[HTML]{D7D7D7}{\color[HTML]{327640} \textbf{82.08}} \\ \hline
                           & Original                         & 72.87                                                         & 94.30                                                         & 37.79                                                         & 83.59                                                         & 90.65                                                         & 98.84                                                         & 9.93                                                          & 94.75                                                         & 65.23                                                         & 93.94                                                         & 18.11                                                         & 79.59                                                         \\
                           & \cellcolor[HTML]{D7D7D7}+LW-FEST & \cellcolor[HTML]{D7D7D7}73.24                                 & \cellcolor[HTML]{D7D7D7}{\color[HTML]{327640} \textbf{96.20}} & \cellcolor[HTML]{D7D7D7}50.83                                 & \cellcolor[HTML]{D7D7D7}84.72                                 & \cellcolor[HTML]{D7D7D7}92.54                                 & \cellcolor[HTML]{D7D7D7}99.15                                 & \cellcolor[HTML]{D7D7D7}{\color[HTML]{327640} \textbf{9.79}}  & \cellcolor[HTML]{D7D7D7}95.85                                 & \cellcolor[HTML]{D7D7D7}65.70                                 & \cellcolor[HTML]{D7D7D7}94.95                                 & \cellcolor[HTML]{D7D7D7}{\color[HTML]{327640} \textbf{16.25}} & \cellcolor[HTML]{D7D7D7}80.33                                 \\
\multirow{-3}{*}{MLCL-Net \textsuperscript{\cite{yu2022infrared}}} & \cellcolor[HTML]{D7D7D7}+FEST    & \cellcolor[HTML]{D7D7D7}{\color[HTML]{327640} \textbf{73.61}} & \cellcolor[HTML]{D7D7D7}{\color[HTML]{327640} \textbf{96.20}} & \cellcolor[HTML]{D7D7D7}{\color[HTML]{327640} \textbf{34.78}} & \cellcolor[HTML]{D7D7D7}{\color[HTML]{327640} \textbf{84.91}} & \cellcolor[HTML]{D7D7D7}{\color[HTML]{327640} \textbf{92.70}} & \cellcolor[HTML]{D7D7D7}{\color[HTML]{327640} \textbf{99.37}} & \cellcolor[HTML]{D7D7D7}10.32                                 & \cellcolor[HTML]{D7D7D7}{\color[HTML]{327640} \textbf{96.04}} & \cellcolor[HTML]{D7D7D7}{\color[HTML]{327640} \textbf{66.20}} & \cellcolor[HTML]{D7D7D7}{\color[HTML]{327640} \textbf{95.62}} & \cellcolor[HTML]{D7D7D7}18.09                                 & \cellcolor[HTML]{D7D7D7}{\color[HTML]{327640} \textbf{80.91}} \\ \hline
                           & Original                         & 75.29                                                         & 95.44                                                         & 36.84                                                         & 85.37                                                         & 91.52                                                         & 98.41                                                         & {\color[HTML]{327640} \textbf{5.84}}                          & 94.97                                                         & 66.91                                                         & 93.94                                                         & 29.68                                                         & 80.43                                                         \\
                           & \cellcolor[HTML]{D7D7D7}+LW-FEST & \cellcolor[HTML]{D7D7D7}75.71                                 & \cellcolor[HTML]{D7D7D7}96.20                                 & \cellcolor[HTML]{D7D7D7}23.94                                 & \cellcolor[HTML]{D7D7D7}85.96                                 & \cellcolor[HTML]{D7D7D7}92.65                                 & \cellcolor[HTML]{D7D7D7}99.47                                 & \cellcolor[HTML]{D7D7D7}10.34                                 & \cellcolor[HTML]{D7D7D7}96.06                                 & \cellcolor[HTML]{D7D7D7}{\color[HTML]{327640} \textbf{67.47}} & \cellcolor[HTML]{D7D7D7}95.96                                 & \cellcolor[HTML]{D7D7D7}34.92                                 & \cellcolor[HTML]{D7D7D7}81.72                                 \\
\multirow{-3}{*}{ALCL-Net \textsuperscript{\cite{yu2022pay}}} & \cellcolor[HTML]{D7D7D7}+FEST    & \cellcolor[HTML]{D7D7D7}{\color[HTML]{327640} \textbf{77.03}} & \cellcolor[HTML]{D7D7D7}{\color[HTML]{327640} \textbf{96.96}} & \cellcolor[HTML]{D7D7D7}{\color[HTML]{327640} \textbf{18.87}} & \cellcolor[HTML]{D7D7D7}{\color[HTML]{327640} \textbf{87.00}} & \cellcolor[HTML]{D7D7D7}{\color[HTML]{327640} \textbf{92.86}} & \cellcolor[HTML]{D7D7D7}{\color[HTML]{327640} \textbf{99.58}} & \cellcolor[HTML]{D7D7D7}6.18                                  & \cellcolor[HTML]{D7D7D7}{\color[HTML]{327640} \textbf{96.22}} & \cellcolor[HTML]{D7D7D7}{\color[HTML]{327640} \textbf{67.47}} & \cellcolor[HTML]{D7D7D7}{\color[HTML]{327640} \textbf{97.64}} & \cellcolor[HTML]{D7D7D7}{\color[HTML]{327640} \textbf{22.72}} & \cellcolor[HTML]{D7D7D7}{\color[HTML]{327640} \textbf{82.56}} \\ \hline
                           & Original                         & 76.25                                                         & 96.20                                                         & 29.56                                                         & 86.23                                                         & 91.14                                                         & 98.73                                                         & 10.50                                                         & 94.94                                                         & 68.69                                                         & 94.61                                                         & 43.49                                                         & 81.65                                                         \\
                           & \cellcolor[HTML]{D7D7D7}+LW-FEST & \cellcolor[HTML]{D7D7D7}77.40                                 & \cellcolor[HTML]{D7D7D7}97.72                                 & \cellcolor[HTML]{D7D7D7}26.34                                 & \cellcolor[HTML]{D7D7D7}87.56                                 & \cellcolor[HTML]{D7D7D7}92.25                                 & \cellcolor[HTML]{D7D7D7}99.37                                 & \cellcolor[HTML]{D7D7D7}18.13                                 & \cellcolor[HTML]{D7D7D7}95.81                                 & \cellcolor[HTML]{D7D7D7}69.43                                 & \cellcolor[HTML]{D7D7D7}95.62                                 & \cellcolor[HTML]{D7D7D7}38.32                                 & \cellcolor[HTML]{D7D7D7}82.53                                 \\
\multirow{-3}{*}{DNA-Net \textsuperscript{\cite{li2022dense}}}  & \cellcolor[HTML]{D7D7D7}+FEST    & \cellcolor[HTML]{D7D7D7}{\color[HTML]{327640} \textbf{79.06}} & \cellcolor[HTML]{D7D7D7}{\color[HTML]{327640} \textbf{98.10}} & \cellcolor[HTML]{D7D7D7}{\color[HTML]{327640} \textbf{24.97}} & \cellcolor[HTML]{D7D7D7}{\color[HTML]{327640} \textbf{88.58}} & \cellcolor[HTML]{D7D7D7}{\color[HTML]{327640} \textbf{94.20}} & \cellcolor[HTML]{D7D7D7}{\color[HTML]{327640} \textbf{99.47}} & \cellcolor[HTML]{D7D7D7}{\color[HTML]{327640} \textbf{6.99}}  & \cellcolor[HTML]{D7D7D7}{\color[HTML]{327640} \textbf{96.84}} & \cellcolor[HTML]{D7D7D7}{\color[HTML]{327640} \textbf{70.09}} & \cellcolor[HTML]{D7D7D7}{\color[HTML]{327640} \textbf{96.30}} & \cellcolor[HTML]{D7D7D7}{\color[HTML]{327640} \textbf{26.97}} & \cellcolor[HTML]{D7D7D7}{\color[HTML]{327640} \textbf{83.20}} \\ \hline
                           & Original                         & 78.77                                                         & 98.10                                                         & 31.53                                                         & 88.44                                                         & 92.91                                                         & 99.15                                                         & {\color[HTML]{327640} \textbf{2.57}}                          & 96.03                                                         & 68.40                                                         & 93.60                                                         & {\color[HTML]{327640} \textbf{17.23}}                         & 81.00                                                         \\
                           & \cellcolor[HTML]{D7D7D7}+LW-FEST & \cellcolor[HTML]{D7D7D7}78.82                                 & \cellcolor[HTML]{D7D7D7}{\color[HTML]{327640} \textbf{98.48}} & \cellcolor[HTML]{D7D7D7}22.43                                 & \cellcolor[HTML]{D7D7D7}88.65                                 & \cellcolor[HTML]{D7D7D7}93.34                                 & \cellcolor[HTML]{D7D7D7}99.47                                 & \cellcolor[HTML]{D7D7D7}3.13                                  & \cellcolor[HTML]{D7D7D7}96.41                                 & \cellcolor[HTML]{D7D7D7}67.77                                 & \cellcolor[HTML]{D7D7D7}94.95                                 & \cellcolor[HTML]{D7D7D7}23.57                                 & \cellcolor[HTML]{D7D7D7}81.36                                 \\
\multirow{-3}{*}{GGL-Net \textsuperscript{\cite{zhao2023gradient}}}  & \cellcolor[HTML]{D7D7D7}+FEST    & \cellcolor[HTML]{D7D7D7}{\color[HTML]{327640} \textbf{79.59}} & \cellcolor[HTML]{D7D7D7}98.10                                 & \cellcolor[HTML]{D7D7D7}{\color[HTML]{327640} \textbf{17.70}} & \cellcolor[HTML]{D7D7D7}{\color[HTML]{327640} \textbf{88.85}} & \cellcolor[HTML]{D7D7D7}{\color[HTML]{327640} \textbf{93.39}} & \cellcolor[HTML]{D7D7D7}{\color[HTML]{327640} \textbf{99.58}} & \cellcolor[HTML]{D7D7D7}3.79                                  & \cellcolor[HTML]{D7D7D7}{\color[HTML]{327640} \textbf{96.49}} & \cellcolor[HTML]{D7D7D7}{\color[HTML]{327640} \textbf{70.02}} & \cellcolor[HTML]{D7D7D7}{\color[HTML]{327640} \textbf{95.29}} & \cellcolor[HTML]{D7D7D7}25.89                                 & \cellcolor[HTML]{D7D7D7}{\color[HTML]{327640} \textbf{82.66}} \\ \hline
                           & Original                         & 77.08                                                         & {\color[HTML]{327640} \textbf{96.20}}                         & 46.50                                                         & 86.64                                                         & 92.04                                                         & 98.31                                                         & 14.02                                                         & 95.18                                                         & 70.70                                                         & 94.26                                                         & 31.79                                                         & 82.48                                                         \\
                           & \cellcolor[HTML]{D7D7D7}+LW-FEST & \cellcolor[HTML]{D7D7D7}77.29                                 & \cellcolor[HTML]{D7D7D7}{\color[HTML]{327640} \textbf{96.20}} & \cellcolor[HTML]{D7D7D7}31.69                                 & \cellcolor[HTML]{D7D7D7}86.75                                 & \cellcolor[HTML]{D7D7D7}93.21                                 & \cellcolor[HTML]{D7D7D7}{\color[HTML]{327640} \textbf{99.58}} & \cellcolor[HTML]{D7D7D7}9.15                                  & \cellcolor[HTML]{D7D7D7}96.40                                 & \cellcolor[HTML]{D7D7D7}72.36                                 & \cellcolor[HTML]{D7D7D7}95.62                                 & \cellcolor[HTML]{D7D7D7}33.04                                 & \cellcolor[HTML]{D7D7D7}83.99                                 \\
\multirow{-3}{*}{UIUNet \textsuperscript{\cite{wu2022uiu}}}   & \cellcolor[HTML]{D7D7D7}+FEST    & \cellcolor[HTML]{D7D7D7}{\color[HTML]{327640} \textbf{79.37}} & \cellcolor[HTML]{D7D7D7}{\color[HTML]{327640} \textbf{96.20}} & \cellcolor[HTML]{D7D7D7}{\color[HTML]{327640} \textbf{29.98}} & \cellcolor[HTML]{D7D7D7}{\color[HTML]{327640} \textbf{87.79}} & \cellcolor[HTML]{D7D7D7}{\color[HTML]{327640} \textbf{93.71}} & \cellcolor[HTML]{D7D7D7}{\color[HTML]{327640} \textbf{99.58}} & \cellcolor[HTML]{D7D7D7}{\color[HTML]{327640} \textbf{4.04}}  & \cellcolor[HTML]{D7D7D7}{\color[HTML]{327640} \textbf{96.65}} & \cellcolor[HTML]{D7D7D7}{\color[HTML]{327640} \textbf{72.86}} & \cellcolor[HTML]{D7D7D7}{\color[HTML]{327640} \textbf{96.97}} & \cellcolor[HTML]{D7D7D7}{\color[HTML]{327640} \textbf{29.57}} & \cellcolor[HTML]{D7D7D7}{\color[HTML]{327640} \textbf{84.92}} \\ \hline
                           & Original                         & 77.43                                                         & 96.20                                                         & 42.10                                                         & 86.82                                                         & 93.21                                                         & 98.73                                                         & 2.94                                                          & 95.97                                                         & 71.03                                                         & 93.94                                                         & {\color[HTML]{327640} \textbf{20.10}}                         & 82.49                                                         \\
                           & \cellcolor[HTML]{D7D7D7}+LW-FEST & \cellcolor[HTML]{D7D7D7}78.27                                 & \cellcolor[HTML]{D7D7D7}{\color[HTML]{327640} \textbf{98.10}} & \cellcolor[HTML]{D7D7D7}20.17                                 & \cellcolor[HTML]{D7D7D7}88.19                                 & \cellcolor[HTML]{D7D7D7}94.29                                 & \cellcolor[HTML]{D7D7D7}99.37                                 & \cellcolor[HTML]{D7D7D7}{\color[HTML]{327640} \textbf{1.82}}  & \cellcolor[HTML]{D7D7D7}96.83                                 & \cellcolor[HTML]{D7D7D7}70.89                                 & \cellcolor[HTML]{D7D7D7}96.00                                 & \cellcolor[HTML]{D7D7D7}28.87                                 & \cellcolor[HTML]{D7D7D7}83.45                                 \\
\multirow{-3}{*}{MSDA-Net \textsuperscript{\cite{zhao2024multi}}} & \cellcolor[HTML]{D7D7D7}+FEST    & \cellcolor[HTML]{D7D7D7}{\color[HTML]{327640} \textbf{80.51}} & \cellcolor[HTML]{D7D7D7}97.72                                 & \cellcolor[HTML]{D7D7D7}{\color[HTML]{327640} \textbf{15.23}} & \cellcolor[HTML]{D7D7D7}{\color[HTML]{327640} \textbf{89.12}} & \cellcolor[HTML]{D7D7D7}{\color[HTML]{327640} \textbf{94.39}} & \cellcolor[HTML]{D7D7D7}{\color[HTML]{327640} \textbf{99.68}} & \cellcolor[HTML]{D7D7D7}3.13                                  & \cellcolor[HTML]{D7D7D7}{\color[HTML]{327640} \textbf{97.04}} & \cellcolor[HTML]{D7D7D7}{\color[HTML]{327640} \textbf{71.19}} & \cellcolor[HTML]{D7D7D7}{\color[HTML]{327640} \textbf{97.98}} & \cellcolor[HTML]{D7D7D7}25.07                                 & \cellcolor[HTML]{D7D7D7}{\color[HTML]{327640} \textbf{84.59}} \\ \hline
\end{tabular}

}}

\end{minipage}
\end{table*}

\vspace{10pt}

\subsection{Discussion}

\hla{\textit{1) Why can the EEDM loss generally and effectively improve the performance of SIRST network?} The proposed EEDM loss is designed for the “small and weak" characteristics of infrared small targets. Its core mechanism consists of two parts: edge pixel enhancement and difficult pixel mining. Specifically, edge pixel enhancement refers to assigning higher weights to the boundary regions of the target, thereby enhancing the model’s sensitivity to target boundaries. Difficult pixel mining refers to discarding a certain proportion of simple samples, so that network training focuses more on difficult areas, thereby prompting the network to learn more discriminative features. The EEDM loss function enables the network to obtain effective supervision signals and achieve effective performance improvement, without requiring any modifications to the network structure or introducing extra overhead during the inference phase.}

\hla{\textit{2) What is the core value of the AS strategy?} The AS strategy breaks through the inherent limitations of single threshold segmentation and fully explores and utilizes the information in the probability mask. Specifically, for SIRST detection scenarios, we introduce the novel concepts of strong targets and weak targets, and adopt a dual threshold mechanism to divide the probability mask into high-confidence strong target regions and relatively low-confidence weak target regions. This strategy preserves the fine segmentation of strong targets while recovering potential missed weak targets at very low cost, thereby improving the overall recall rate and detection performance. In addition, this strategy exhibits good generality and can be seamlessly integrated into existing networks.}

\begin{table*}[]
\centering
\caption{\hla{Performance comparison of the original networks and their LW-FEST and FEST variants on the IRSTD-1k dataset across different difficulty subsets (easy, medium, and hard). $D$ denotes composite difficulty of a sample. $Num$ denotes the number of samples in each subset.  Green denotes the best result within each group, where each group comprises a method’s Original, +LW-FEST, and +FEST variants.}}
\vspace{-5pt}
\label{tab:tab_difficultg_subset_results}
\renewcommand{\arraystretch}{1.1}
\setlength{\tabcolsep}{3mm}
\resizebox{\textwidth}{!}{
\begin{tabular}{c|c|cccc|cccc|cccc}
\hline
                           & & \multicolumn{4}{c|}{Easy ($D \leq 0.3$), $Num$=2} 
& \multicolumn{4}{c|}{Medium ($0.3 < D < 0.7$), $Num$=102} 
& \multicolumn{4}{c}{Hard ($D \geq 0.7$), $Num$=97} \\ \cline{3-14}                                                               
\multirow{-2}{*}{Scheme}   & \multirow{-2}{*}{Description}     & \textit{IoU} ↑                                                   & \textit{$P_d$} ↑                                                    & \textit{$F_a$}  ↓                                                  & \textit{Score} ↑                                                 & \textit{IoU} ↑                                                 & \textit{$P_d$} ↑                                                   & \textit{$F_a$}  ↓                                                   & \textit{Score} ↑                                                & \textit{IoU} ↑                                                  & \textit{$P_d$} ↑                                                   & \textit{$F_a$}  ↓                                                   & \textit{Score} ↑                                                \\ \hline
                           & Original                          & 94.44                                                          & {\color[HTML]{327640} \textbf{100.00}}                         & {\color[HTML]{327640} \textbf{0.00}}                         & 97.22                                                          & 72.87                                                         & 95.24                                                         & 6.28                                                         & 84.06                                                         & 65.26                                                         & 93.92                                                         & 83.25                                                         & 79.59                                                         \\
                           & \cellcolor[HTML]{D7D7D7}+ LW-FEST & \cellcolor[HTML]{D7D7D7}{\color[HTML]{327640} \textbf{100.00}} & \cellcolor[HTML]{D7D7D7}{\color[HTML]{327640} \textbf{100.00}}  & \cellcolor[HTML]{D7D7D7}{\color[HTML]{327640} \textbf{0.00}} & \cellcolor[HTML]{D7D7D7}{\color[HTML]{327640} \textbf{100.00}} & \cellcolor[HTML]{D7D7D7}73.46                                 & \cellcolor[HTML]{D7D7D7}{\color[HTML]{327640} \textbf{96.60}} & \cellcolor[HTML]{D7D7D7}6.17                                 & \cellcolor[HTML]{D7D7D7}85.03                                 & \cellcolor[HTML]{D7D7D7}66.06                                 & \cellcolor[HTML]{D7D7D7}94.59                                 & \cellcolor[HTML]{D7D7D7}72.91                                 & \cellcolor[HTML]{D7D7D7}80.33                                 \\
\multirow{-3}{*}{DNA-Net}  & \cellcolor[HTML]{D7D7D7}+FEST     & \cellcolor[HTML]{D7D7D7}{\color[HTML]{327640} \textbf{100.00}} & \cellcolor[HTML]{D7D7D7}{\color[HTML]{327640} \textbf{100.00}}  & \cellcolor[HTML]{D7D7D7}{\color[HTML]{327640} \textbf{0.00}} & \cellcolor[HTML]{D7D7D7}{\color[HTML]{327640} \textbf{100.00}} & \cellcolor[HTML]{D7D7D7}{\color[HTML]{327640} \textbf{73.48}} & \cellcolor[HTML]{D7D7D7}{\color[HTML]{327640} \textbf{96.60}} & \cellcolor[HTML]{D7D7D7}{\color[HTML]{327640} \textbf{4.97}} & \cellcolor[HTML]{D7D7D7}{\color[HTML]{327640} \textbf{85.04}} & \cellcolor[HTML]{D7D7D7}{\color[HTML]{327640} \textbf{67.10}} & \cellcolor[HTML]{D7D7D7}{\color[HTML]{327640} \textbf{95.95}} & \cellcolor[HTML]{D7D7D7}{\color[HTML]{327640} \textbf{50.65}} & \cellcolor[HTML]{D7D7D7}{\color[HTML]{327640} \textbf{81.53}} \\ \hline
                           & Original                          & {\color[HTML]{327640} \textbf{94.74}}                          & {\color[HTML]{327640} \textbf{100.00}}                         & {\color[HTML]{327640} \textbf{0.00}}                         & {\color[HTML]{327640} \textbf{97.37}}                          & {\color[HTML]{327640} \textbf{74.77}}                         & 94.56                                                         & 5.12                                                         & 84.67                                                         & 62.81                                                         & 90.54                                                         & 30.32                                                         & 76.68                                                         \\
                           & \cellcolor[HTML]{D7D7D7}+ LW-FEST & \cellcolor[HTML]{D7D7D7}90.00                                  & \cellcolor[HTML]{D7D7D7}{\color[HTML]{327640} \textbf{100.00}} & \cellcolor[HTML]{D7D7D7}{\color[HTML]{327640} \textbf{0.00}} & \cellcolor[HTML]{D7D7D7}95.00                                  & \cellcolor[HTML]{D7D7D7}74.13                                 & \cellcolor[HTML]{D7D7D7}{\color[HTML]{327640} \textbf{96.60}} & \cellcolor[HTML]{D7D7D7}{\color[HTML]{327640} \textbf{4.08}} & \cellcolor[HTML]{D7D7D7}85.37                                 & \cellcolor[HTML]{D7D7D7}62.60                                 & \cellcolor[HTML]{D7D7D7}93.24                                 & \cellcolor[HTML]{D7D7D7}49.36                                 & \cellcolor[HTML]{D7D7D7}77.92                                 \\
\multirow{-3}{*}{GGL-Net}  & \cellcolor[HTML]{D7D7D7}+FEST     & \cellcolor[HTML]{D7D7D7}{\color[HTML]{327640} \textbf{94.74}}  & \cellcolor[HTML]{D7D7D7}{\color[HTML]{327640} \textbf{100.00}} & \cellcolor[HTML]{D7D7D7}{\color[HTML]{327640} \textbf{0.00}} & \cellcolor[HTML]{D7D7D7}{\color[HTML]{327640} \textbf{97.37}}  & \cellcolor[HTML]{D7D7D7}74.32                                 & \cellcolor[HTML]{D7D7D7}{\color[HTML]{327640} \textbf{96.60}} & \cellcolor[HTML]{D7D7D7}4.30                                 & \cellcolor[HTML]{D7D7D7}{\color[HTML]{327640} \textbf{85.46}} & \cellcolor[HTML]{D7D7D7}{\color[HTML]{327640} \textbf{66.25}} & \cellcolor[HTML]{D7D7D7}{\color[HTML]{327640} \textbf{93.92}} & \cellcolor[HTML]{D7D7D7}{\color[HTML]{327640} \textbf{49.12}} & \cellcolor[HTML]{D7D7D7}{\color[HTML]{327640} \textbf{80.09}} \\ \hline
                           & Original                          & 90.00                                                          & {\color[HTML]{327640} \textbf{100.00}}                         & {\color[HTML]{327640} \textbf{0.00}}                         & 95.00                                                          & 73.58                                                         & 95.24                                                         & 22.10                                                        & 84.41                                                         & 68.36                                                         & 93.24                                                         & {\color[HTML]{327640} \textbf{42.63}}                         & 80.80                                                         \\
                           & \cellcolor[HTML]{D7D7D7}+ LW-FEST & \cellcolor[HTML]{D7D7D7}{\color[HTML]{327640} \textbf{100.00}} & \cellcolor[HTML]{D7D7D7}{\color[HTML]{327640} \textbf{100.00}} & \cellcolor[HTML]{D7D7D7}{\color[HTML]{327640} \textbf{0.00}} & \cellcolor[HTML]{D7D7D7}{\color[HTML]{327640} \textbf{100.00}} & \cellcolor[HTML]{D7D7D7}{\color[HTML]{327640} \textbf{75.84}} & \cellcolor[HTML]{D7D7D7}97.28                                 & \cellcolor[HTML]{D7D7D7}4.30                                 & \cellcolor[HTML]{D7D7D7}{\color[HTML]{327640} \textbf{86.56}} & \cellcolor[HTML]{D7D7D7}69.45                                 & \cellcolor[HTML]{D7D7D7}93.92                                 & \cellcolor[HTML]{D7D7D7}63.95                                 & \cellcolor[HTML]{D7D7D7}81.69                                 \\
\multirow{-3}{*}{UIU-Net}  & \cellcolor[HTML]{D7D7D7}+FEST     & \cellcolor[HTML]{D7D7D7}{\color[HTML]{327640} \textbf{100.00}} & \cellcolor[HTML]{D7D7D7}{\color[HTML]{327640} \textbf{100.00}} & \cellcolor[HTML]{D7D7D7}{\color[HTML]{327640} \textbf{0.00}} & \cellcolor[HTML]{D7D7D7}{\color[HTML]{327640} \textbf{100.00}} & \cellcolor[HTML]{D7D7D7}74.86                                 & \cellcolor[HTML]{D7D7D7}{\color[HTML]{327640} \textbf{97.96}} & \cellcolor[HTML]{D7D7D7}{\color[HTML]{327640} \textbf{3.18}} & \cellcolor[HTML]{D7D7D7}86.41                                 & \cellcolor[HTML]{D7D7D7}{\color[HTML]{327640} \textbf{71.11}} & \cellcolor[HTML]{D7D7D7}{\color[HTML]{327640} \textbf{95.95}} & \cellcolor[HTML]{D7D7D7}57.93                                 & \cellcolor[HTML]{D7D7D7}{\color[HTML]{327640} \textbf{83.53}} \\ \hline
                           & Original                          & {\color[HTML]{327640} \textbf{100.00}}                         & {\color[HTML]{327640} \textbf{100.00}}                         & {\color[HTML]{327640} \textbf{0.00}}                         & {\color[HTML]{327640} \textbf{100.00}}                         & {\color[HTML]{327640} \textbf{73.75}}                         & 96.60                                                         & 9.01                                                         & 85.18                                                         & 68.68                                                         & 91.21                                                         & 32.17                                                         & 79.95                                                         \\
                           & \cellcolor[HTML]{D7D7D7}+ LW-FEST & \cellcolor[HTML]{D7D7D7}{\color[HTML]{327640} \textbf{100.00}} & \cellcolor[HTML]{D7D7D7}{\color[HTML]{327640} \textbf{100.00}} & \cellcolor[HTML]{D7D7D7}{\color[HTML]{327640} \textbf{0.00}} & \cellcolor[HTML]{D7D7D7}{\color[HTML]{327640} \textbf{100.00}} & \cellcolor[HTML]{D7D7D7}73.32                                 & \cellcolor[HTML]{D7D7D7}97.28                                 & \cellcolor[HTML]{D7D7D7}5.83                                 & \cellcolor[HTML]{D7D7D7}85.30                                 & \cellcolor[HTML]{D7D7D7}68.78                                 & \cellcolor[HTML]{D7D7D7}94.59                                 & \cellcolor[HTML]{D7D7D7}53.68                                 & \cellcolor[HTML]{D7D7D7}81.69                                 \\
\multirow{-3}{*}{MSDA-Net} & \cellcolor[HTML]{D7D7D7}+FEST     & \cellcolor[HTML]{D7D7D7}{\color[HTML]{327640} \textbf{100.00}} & \cellcolor[HTML]{D7D7D7}{\color[HTML]{327640} \textbf{100.00}} & \cellcolor[HTML]{D7D7D7}{\color[HTML]{327640} \textbf{0.00}} & \cellcolor[HTML]{D7D7D7}{\color[HTML]{327640} \textbf{100.00}} & \cellcolor[HTML]{D7D7D7}73.56                                 & \cellcolor[HTML]{D7D7D7}{\color[HTML]{327640} \textbf{97.96}} & \cellcolor[HTML]{D7D7D7}{\color[HTML]{327640} \textbf{4.00}} & \cellcolor[HTML]{D7D7D7}{\color[HTML]{327640} \textbf{85.76}} & \cellcolor[HTML]{D7D7D7}{\color[HTML]{327640} \textbf{69.11}} & \cellcolor[HTML]{D7D7D7}{\color[HTML]{327640} \textbf{97.97}} & \cellcolor[HTML]{D7D7D7}{\color[HTML]{327640} \textbf{47.74}} & \cellcolor[HTML]{D7D7D7}{\color[HTML]{327640} \textbf{83.54}} \\ \hline
\end{tabular}
}
\vspace{3pt}
\end{table*}

\begin{table*}[t]
\centering
\caption{\hla{Performance comparison of different networks before and after using LW-FEST and FEST frameworks on the IRSTD-1k dataset. (Params: M, Time: ms).}}
\vspace{-5pt}
\label{tab:tab_cal_time_gflops}
\renewcommand{\arraystretch}{1.4}
\setlength{\tabcolsep}{2.5mm}{
\resizebox{\textwidth}{!}{
\begin{tabular}{c|cccc|cccc|cccc}
\hline
{\color[HTML]{000000} }                         & \multicolumn{4}{c|}{{\color[HTML]{000000} Original}}                                                                                                                                                         & \multicolumn{4}{c|}{{\color[HTML]{000000} Original + LW-FEST}}                                                                                                                                                                & \multicolumn{4}{c}{{\color[HTML]{000000} Original + FEST}}                                                                                                                                                      \\ \cline{2-13} 
\multirow{-2}{*}{{\color[HTML]{000000} Method}} & \multicolumn{1}{c|}{{\color[HTML]{000000} \textit{\textbf{Score}}}} & \multicolumn{1}{c|}{{\color[HTML]{000000} Params}} & \multicolumn{1}{c|}{{\color[HTML]{000000} GFLOPs}} & {\color[HTML]{000000} Time}  & \multicolumn{1}{c|}{{\color[HTML]{000000} \textit{\textbf{Score}}}} & \multicolumn{1}{c|}{{\color[HTML]{000000} Params}}         & \multicolumn{1}{c|}{{\color[HTML]{000000} GFLOPs}}          & {\color[HTML]{000000} Time}  & \multicolumn{1}{c|}{{\color[HTML]{000000} \textit{\textbf{Score}}}}  & \multicolumn{1}{c|}{{\color[HTML]{000000} Params}} & \multicolumn{1}{c|}{{\color[HTML]{000000} GFLOPs}}  & {\color[HTML]{000000} Time}   \\ \hline
{\color[HTML]{000000} ACM}                      & \multicolumn{1}{c|}{{\color[HTML]{000000} 74.93}}                   & \multicolumn{1}{c|}{{\color[HTML]{000000} 0.40}}   & \multicolumn{1}{c|}{{\color[HTML]{000000} 3.24}}   & {\color[HTML]{000000} 1.97}  & \multicolumn{1}{c|}{{\color[HTML]{000000} 79.02 (↑5.46\%)}}         & \multicolumn{1}{c|}{{\color[HTML]{000000} \textbf{0.40}}}  & \multicolumn{1}{c|}{{\color[HTML]{000000} \textbf{3.24}}}   & {\color[HTML]{000000} 2.00}  & \multicolumn{1}{c|}{{\color[HTML]{000000} \textbf{80.37 (↑7.26\%)}}} & \multicolumn{1}{c|}{{\color[HTML]{000000} 0.80}}   & \multicolumn{1}{c|}{{\color[HTML]{000000} 10.54}}   & {\color[HTML]{000000} 3.97}   \\ \hline
{\color[HTML]{000000} ALCNet}                   & \multicolumn{1}{c|}{{\color[HTML]{000000} 77.76}}                   & \multicolumn{1}{c|}{{\color[HTML]{000000} 0.43}}   & \multicolumn{1}{c|}{{\color[HTML]{000000} 3.05}}   & {\color[HTML]{000000} 1.94}  & \multicolumn{1}{c|}{{\color[HTML]{000000} 79.94 (↑2.80\%)}}         & \multicolumn{1}{c|}{{\color[HTML]{000000} \textbf{0.43}}}  & \multicolumn{1}{c|}{{\color[HTML]{000000} \textbf{3.05}}}   & {\color[HTML]{000000} 1.99}  & \multicolumn{1}{c|}{{\color[HTML]{000000} \textbf{82.08 (↑5.56\%)}}} & \multicolumn{1}{c|}{{\color[HTML]{000000} 0.85}}   & \multicolumn{1}{c|}{{\color[HTML]{000000} 9.90}}    & {\color[HTML]{000000} 3.91}   \\ \hline
{\color[HTML]{000000} MLCL-Net}                 & \multicolumn{1}{c|}{{\color[HTML]{000000} 79.59}}                   & \multicolumn{1}{c|}{{\color[HTML]{000000} 0.56}}   & \multicolumn{1}{c|}{{\color[HTML]{000000} 49.71}}  & {\color[HTML]{000000} 4.20}  & \multicolumn{1}{c|}{{\color[HTML]{000000} 80.33 (↑0.93\%)}}         & \multicolumn{1}{c|}{{\color[HTML]{000000} \textbf{0.56}}}  & \multicolumn{1}{c|}{{\color[HTML]{000000} \textbf{49.71}}}  & {\color[HTML]{000000} 4.23}  & \multicolumn{1}{c|}{{\color[HTML]{000000} \textbf{80.91 (↑1.66\%)}}} & \multicolumn{1}{c|}{{\color[HTML]{000000} 1.12}}   & \multicolumn{1}{c|}{{\color[HTML]{000000} 161.56}}  & {\color[HTML]{000000} 8.44}   \\ \hline
{\color[HTML]{000000} ALCL-Net}                 & \multicolumn{1}{c|}{{\color[HTML]{000000} 80.43}}                   & \multicolumn{1}{c|}{{\color[HTML]{000000} 5.67}}   & \multicolumn{1}{c|}{{\color[HTML]{000000} 54.44}}  & {\color[HTML]{000000} 4.12}  & \multicolumn{1}{c|}{{\color[HTML]{000000} 81.72 (↑1.60\%)}}         & \multicolumn{1}{c|}{{\color[HTML]{000000} \textbf{5.67}}}  & \multicolumn{1}{c|}{{\color[HTML]{000000} \textbf{54.44}}}  & {\color[HTML]{000000} 4.14}  & \multicolumn{1}{c|}{{\color[HTML]{000000} \textbf{82.56 (↑2.65\%)}}} & \multicolumn{1}{c|}{{\color[HTML]{000000} 11.34}}  & \multicolumn{1}{c|}{{\color[HTML]{000000} 176.93}}  & {\color[HTML]{000000} 8.28}   \\ \hline
{\color[HTML]{000000} DNA-Net}                  & \multicolumn{1}{c|}{{\color[HTML]{000000} 81.65}}                   & \multicolumn{1}{c|}{{\color[HTML]{000000} 4.70}}   & \multicolumn{1}{c|}{{\color[HTML]{000000} 114.26}} & {\color[HTML]{000000} 67.03} & \multicolumn{1}{c|}{{\color[HTML]{000000} 82.53 (↑1.08\%)}}         & \multicolumn{1}{c|}{{\color[HTML]{000000} \textbf{4.70}}}  & \multicolumn{1}{c|}{{\color[HTML]{000000} \textbf{114.26}}} & {\color[HTML]{000000} 67.04} & \multicolumn{1}{c|}{{\color[HTML]{000000} \textbf{83.20 (↑1.90\%)}}} & \multicolumn{1}{c|}{{\color[HTML]{000000} 9.39}}   & \multicolumn{1}{c|}{{\color[HTML]{000000} 371.34}}  & {\color[HTML]{000000} 133.98} \\ \hline
{\color[HTML]{000000} GGL-Net}                  & \multicolumn{1}{c|}{{\color[HTML]{000000} 81.00}}                   & \multicolumn{1}{c|}{{\color[HTML]{000000} 8.99}}   & \multicolumn{1}{c|}{{\color[HTML]{000000} 78.85}}  & {\color[HTML]{000000} 15.24} & \multicolumn{1}{c|}{{\color[HTML]{000000} 81.36 (↑0.44\%)}}         & \multicolumn{1}{c|}{{\color[HTML]{000000} \textbf{8.99}}}  & \multicolumn{1}{c|}{{\color[HTML]{000000} \textbf{78.85}}}  & {\color[HTML]{000000} 15.26} & \multicolumn{1}{c|}{{\color[HTML]{000000} \textbf{82.66 (↑2.05\%)}}} & \multicolumn{1}{c|}{{\color[HTML]{000000} 17.98}}  & \multicolumn{1}{c|}{{\color[HTML]{000000} 256.25}}  & {\color[HTML]{000000} 30.53}  \\ \hline
{\color[HTML]{000000} UIUNet}                   & \multicolumn{1}{c|}{{\color[HTML]{000000} 82.48}}                   & \multicolumn{1}{c|}{{\color[HTML]{000000} 50.54}}  & \multicolumn{1}{c|}{{\color[HTML]{000000} 436.01}} & {\color[HTML]{000000} 16.73} & \multicolumn{1}{c|}{{\color[HTML]{000000} 83.99 (↑1.83\%)}}         & \multicolumn{1}{c|}{{\color[HTML]{000000} \textbf{50.54}}} & \multicolumn{1}{c|}{{\color[HTML]{000000} \textbf{436.01}}} & {\color[HTML]{000000} 16.75} & \multicolumn{1}{c|}{{\color[HTML]{000000} \textbf{84.92 (↑2.96\%)}}} & \multicolumn{1}{c|}{{\color[HTML]{000000} 101.08}} & \multicolumn{1}{c|}{{\color[HTML]{000000} 1417.02}} & {\color[HTML]{000000} 33.47}  \\ \hline
{\color[HTML]{000000} MSDA-Net}                 & \multicolumn{1}{c|}{{\color[HTML]{000000} 82.49}}                   & \multicolumn{1}{c|}{{\color[HTML]{000000} 4.81}}   & \multicolumn{1}{c|}{{\color[HTML]{000000} 169.80}} & {\color[HTML]{000000} 16.40} & \multicolumn{1}{c|}{{\color[HTML]{000000} 83.45 (↑1.16\%)}}         & \multicolumn{1}{c|}{{\color[HTML]{000000} \textbf{4.81}}}  & \multicolumn{1}{c|}{{\color[HTML]{000000} \textbf{169.80}}} & {\color[HTML]{000000} 16.42} & \multicolumn{1}{c|}{{\color[HTML]{000000} \textbf{84.69 (↑2.67\%)}}} & \multicolumn{1}{c|}{{\color[HTML]{000000} 9.62}}   & \multicolumn{1}{c|}{{\color[HTML]{000000} 551.85}}  & {\color[HTML]{000000} 32.83}  \\ \hline
\end{tabular}
}}
\end{table*}

\hla{\textit{3) How does the proposed framework perform on samples of different difficulty levels?} To objectively evaluate the detection difficulty of each sample, we define a composite difficulty. It comprises three factors: target size, background complexity, and target-background contrast.}

\hla{First, to quantify the target size, we compute the pixel area from the target mask, take the logarithm of this area as the scale measure, and apply min-max normalization across the dataset. Considering that smaller targets are generally harder to detect, the size difficulty is defined as:}
\begin{gather}
D_t^{size} = 1 - Norm(\log \left| {{F_t}} \right|)
\end{gather}
\hla{where $F_t$ denotes the pixel set of target $t$; $\left| {{F_t}} \right|$ denotes its area; $Norm( \cdot )$ denotes the normalization operator; and $D_t^{size}$ denotes the size difficulty of target $t$.}

\hla{Second, to quantify background complexity, we construct a background ring region ${R_t}$ around the target mask via morphological dilation. The ring has a width of $w$ pixels, and any other target regions within it are excluded. In ${R_t}$, we compute gradient energy to quantify edge and texture complexity and local entropy to quantify intensity uncertainty. The gradient energy is defined as:}
\begin{gather}
G_t = \frac{\sum\limits_{i \in R_t} \left( (\partial_{x}L(i))^2 + (\partial_{y}L(i))^2 \right)}{|R_t|}
\end{gather}
\hla{where ${R_t}$ denotes the background ring of target $t$; $\left| {{R_t}} \right|$ denotes its area; $L(i) \in [0,1]$ denotes the normalized intensity at pixel $i \in {R_t}$; ${\partial _{x}}L(i)$ and ${\partial _{y}}L(i)$ denote the first-order spatial derivatives of the intensity field along the $x$ and $y$ directions, respectively; and $G_t$ denotes the gradient energy. The local entropy is defined as:}
\begin{gather}
{H_t} =  - \sum\limits_{k = 1}^{{B_t}} {{p_k}log({p_k})} 
\end{gather}
\hla{where ${B_t} = \sqrt {\left| {{R_t}} \right|}$ denotes the number of histogram bins in ${R_t}$; ${p_k}$ denotes the probability of the ${k_{th}}$ bin; and ${H_t}$ denotes the local entropy. Taking both measures into account, the background complexity difficulty is defined as:}
\begin{gather}
D_t^{clutter} = max(Norm({G_t}),Norm({H_t}))
\end{gather}
\hla{where $\max ( \cdot )$ denotes the maximum operator; and $D_t^{clutter}$ denotes the background complexity difficulty of the target $t$.}

\hla{Third, to quantify the target-background contrast, we compute the Signal-to-Clutter Ratio (SCR) using the background ring region ${R_t}$. The formula of SCR is as follows:}
\begin{gather}
SC{R_t} = \frac{{{\mu _t} - {\mu _{{R_t}}}}}{{{\sigma _{{R_t}}} + \varepsilon }}
\end{gather}
\hla{where ${\mu _t}$ denotes the mean intensity over the target region; ${\mu_{{R_t}}}$ and ${\sigma_{{R_t}}}$ denote the mean and standard deviation within ${R_t}$; and $\varepsilon$ denotes a small positive constant. The contrast difficulty is defined as:}
\begin{gather}
D_t^{contrast} = 1 - Norm(SC{R_t})
\end{gather}
\hla{where $D_t^{contrast}$ denotes the contrast difficulty of target $t$.}

\hla{Finally, we define the target-level composite difficulty that integrates the three factors above to assess a target’s detection difficulty. The formula is as follows:}
\begin{gather}
{D_{t}} = \alpha D_t^{size} + \beta D_t^{clutter} + \gamma D_{t}^{contrast}
\end{gather}
\hla{where $\alpha$, $\beta$ and $\gamma$ denote weighting coefficients, which are set to 0.3, 0.3, and 0.4 to balance the contributions of the three factors.}

\hla{In addition, we define the image-level composite difficulty as follows: If an image contains a single target, the image difficulty equals that target’s difficulty; if multiple targets are present, the image difficulty is defined as the maximum difficulty among all the targets in the image.}

\hla{To thoroughly evaluate the proposed framework across different difficulty levels (easy, medium, and hard), we stratified the IRSTD-1k dataset into three subsets based on the aforementioned image-level composite difficulty metric and conducted evaluations on each subset. The detailed results are shown in \Cref{tab:tab_difficultg_subset_results}. The experimental results show that the proposed framework achieves good detection performance across all difficulty levels. In particular, the improvement on the hard subset is more significant.}

\hla{\textit{4) What are the respective computational costs of the FEST and LW-FEST frameworks?} To comprehensively evaluate the computational costs and application scenarios of the FEST and LW-FEST frameworks, we conducted comparative experiments on the IRSTD-1k dataset. Specifically, these experiments analyze changes in the Parameters, GFLOPs, and per-image inference time, before and after the introduction of the framework. The detailed results are shown in \Cref{tab:tab_cal_time_gflops}. The experimental results demonstrate that after introducing the LW-FEST framework, the number of parameters and GFLOPs remain unchanged, the inference time only increases slightly, but the performance is steadily improved. Notably, compared with the original SIRST network, the SIRST network equipped with the LW-FEST framework achieves a performance improvement of 0.44\% - 5.46\% in the evaluation metric \textbf{\textit{Score}}. Therefore, the LW-FEST is particularly suitable for application scenarios with high efficiency requirements. At the same time, introducing the accuracy-oriented FEST framework leads to a relatively significant increase in parameters, GFLOPs, and inference time, while bringing more obvious performance improvements. Specifically, compared with the original SIRST network, the SIRST network equipped with the FEST framework achieves a performance improvement of 1.66\% - 7.26\% in the evaluation metric \textbf{\textit{Score}}. Therefore, the FEST is more suitable for application scenarios that prioritize accuracy. Overall, LW-FEST enhances efficiency with negligible computational costs, while FEST further pushes through the performance ceiling through multi-scale fusion, but requires greater resource consumption as a trade-off. These complementary characteristics enable FEST and LW-FEST to meet the requirements of different application scenarios.}

\section{Conclusion}
\label{sec:conclusion}
This manuscript proposes a feature-enhanced and sensitivity-tunable (FEST) framework, aiming to provide an efficient and universal optimization framework for \hla{SIRST detection tasks.} Specifically, we propose an edge enhancement difficulty mining (EEDM) loss, \hla{which guides the network to focus more on challenging regions and edge features,} thereby promoting fine-grained learning and positive optimization of the model. Meanwhile, we adopt a multi-scale fusion strategy, which enhances the \hla{model’s perception to multi-scale features of multi-size targets.} In addition, to fully utilize the probability mask information, we design an adjustable sensitivity (AS) strategy for network post-processing. This strategy employs a tunable dual threshold mechanism, which significantly improves the detection rate of infrared small targets while maintaining the segmentation accuracy of the model. Extensive experimental results demonstrate that the proposed FEST framework achieves significant performance gains and exhibits strong robustness on multiple public datasets. Notably, the lightweight LW-FEST framework achieves notable performance gains with minimal additional resource consumption.

\hla{In this study, although we provide a lightweight LW-FEST, the higher-accuracy FEST still inevitably brings certain computational costs because it employs a multi-scale fusion strategy. In future work, we will develop an adaptive detector in combination with a difficulty prediction mechanism. The detector will maintain lightweight inference on easy samples and dynamically activate more complex modules for hard samples, thereby reducing the overall inference burden. In addition, we plan to extend the current single-frame detection framework to multi-frame detection tasks to fully exploit temporal information and further improve detection performance. Notably, in practical scenarios, small targets often exhibit gradually emerging dynamics. Therefore, we will explore incorporating LSTM and other temporal prediction techniques \cite{sultan2025intelligent,raja2025design,raja2025design2} to combine detection with forecasting of the target’s future position and shape, thereby providing decision support for early warning, interception, and observation-resource scheduling.}

\bibliographystyle{elsarticle-num} 
\bibliography{main}

\end{document}